\documentclass[sigconf]{acmart}
\AtBeginDocument{%
  }

\setcopyright{acmlicensed}
\copyrightyear{2026}
\acmYear{2026}
\acmDOI{XXXXXXX.XXXXXXX}
\acmConference[ACM CAIS 2026]{ACM Conference on AI and Agentic Systems}{May 26--29,
  2026}{San Jose, CA}
\acmISBN{978-1-4503-XXXX-X/2018/06}

\renewcommand\footnotetextcopyrightpermission[1]{}    

\usepackage{booktabs}
\usepackage{makecell}
\usepackage{tabularx}

\usepackage{enumitem}

\usepackage{algorithm}
\usepackage{algpseudocode}

\usepackage{amsmath}

\usepackage[dvipsnames,table]{xcolor}

\newif\ifrevisionmode
\revisionmodefalse
\definecolor{revcolor}{rgb}{0.0,0.0,0.75}
\newcommand{\added}[1]{\ifrevisionmode\textcolor{revcolor}{#1}\else#1\fi}
\newcommand{\rc}[1]{\ifrevisionmode\textsuperscript{\textcolor{revcolor!60!black}{\tiny[\,#1\,]}}\fi}

\usepackage{listings}
\lstdefinelanguage{yaml}{
  keywords={true,false,null,y,n},
  sensitive=false,
  comment=[l]{\#},
  morestring=[b]',
  morestring=[b]",
}
\lstdefinestyle{yaml}{
  language=yaml,
  basicstyle=\footnotesize\ttfamily,
  keywordstyle=\color{blue},
  commentstyle=\color{gray},
  stringstyle=\color{OliveGreen},
  breaklines=true,
  frame=single,
  framerule=0.5pt,
  rulecolor=\color{gray!50},
  backgroundcolor=\color{gray!5},
}
\usepackage[most]{tcolorbox}

\newcommand{\tbl}[1]{\texttt{#1}}      
\newcommand{\col}[1]{\texttt{\textcolor{teal!80!black}{#1}}}      

\newcommand{\ClaudeSDK}{Claude SDK}
\newcommand{\OpenAIAgents}{OpenAI Agents}
\newcommand{\GoogleADK}{Google ADK}
\newcommand{\PydanticAI}{Pydantic AI}
\newcommand{\Agno}{Agno}
\newcommand{\LangChain}{LangChain}
\newcommand{\DSPy}{DSPy}
\newcommand{\Microsoft}{Microsoft}
\newcommand{\Smolagents}{Smolagents}
\newcommand{\AgentScope}{AgentScope}

\newcommand{\badgewidth}{3.2em}
\newcommand{\algbadge}[3]{\null\hfill\makebox[\badgewidth][r]{\scriptsize\fcolorbox{#1}{#2}{\textcolor{#1}{#3}}}}
\newcommand{\llmbadge}{\algbadge{violet}{violet!15}{ReAct}}
\newcommand{\toolbadge}{\algbadge{OliveGreen}{green!15}{tool}}
\newcommand{\atoolbadge}{\algbadge{TealBlue}{TealBlue!15}{Agent-as-a-tool}}
\newcommand{\llmtoolbadge}{\algbadge{OliveGreen}{green!15}{LLM-tool}}

\newcommand{\promptbox}[2]{%
  \begin{tcolorbox}[
    enhanced,
    colback=violet!15,
    colframe=violet!50,
    boxrule=0.5pt,
    rounded corners,
    left=2mm,
    right=2mm,
    top=4mm,
    bottom=1mm,
    attach boxed title to top center={yshift=-\tcboxedtitleheight/2},
    boxed title style={
      colback=TealBlue!15,
      colframe=TealBlue,
      boxrule=0.5pt,
      rounded corners,
    },
    fonttitle=\bfseries\ttfamily\small\color{TealBlue},
    title={#1},
  ]
  #2
  \end{tcolorbox}%
}

\newcommand{\llmtoolbox}[2]{%
  \begin{tcolorbox}[
    enhanced,
    colback=gray!5,
    colframe=gray!50,
    boxrule=0.5pt,
    rounded corners,
    left=2mm,
    right=2mm,
    top=4mm,
    bottom=1mm,
    attach boxed title to top center={yshift=-\tcboxedtitleheight/2},
    boxed title style={
      colback=green!15,
      colframe=OliveGreen,
      boxrule=0.5pt,
      rounded corners,
    },
    fonttitle=\bfseries\ttfamily\small\color{OliveGreen},
    title={#1},
  ]
  #2
  \end{tcolorbox}%
}

\newcommand{\orchestratorbox}[2]{%
  \begin{tcolorbox}[
    enhanced,
    colback=violet!15,
    colframe=violet!50,
    boxrule=0.5pt,
    rounded corners,
    left=2mm,
    right=2mm,
    top=4mm,
    bottom=1mm,
    attach boxed title to top center={yshift=-\tcboxedtitleheight/2},
    boxed title style={
      colback=violet!15,
      colframe=violet,
      boxrule=0.5pt,
      rounded corners,
    },
    fonttitle=\bfseries\ttfamily\small\color{violet},
    title={#1},
  ]
  #2
  \end{tcolorbox}%
}

\usepackage{fontawesome5}

\usepackage{tcolorbox}
\usepackage{tikz}
\usepackage{url}
\usepackage{dirtree}
\usetikzlibrary{shapes.geometric, arrows.meta, positioning, fit, backgrounds, decorations.pathreplacing}
\newcommand{\lock}{\textcolor{gray}{\faLock}}
\newcommand{\pencil}{\textcolor{blue}{\faPencil*}}

\algrenewcommand{\algorithmiccomment}[1]{\hfill\textit{\small\textcolor{gray}{$\triangleright$ #1}}}

\algnewcommand\ParFor{\textbf{parallel for}}
\algnewcommand\EndParFor{\textbf{end parallel for}}
\algdef{SE}[PARFOR]{ParFor}{EndParFor}[1]{\textbf{parallel for} #1 \textbf{do}}{\textbf{end parallel for}}

\begin{document}

\title[Declarative by Design, Assistable Only by Convention]{Declarative by Design, Assistable Only by Convention: Benchmarking Multi-Agent Frameworks for AI-Assistability}

\author{Shafiuddin Rehan Ahmed}
\affiliation{%
  \institution{Center for Advanced AI, Accenture}
  \city{Mountain View}
  \state{California}
  \country{USA}
}
\email{shafiuddin.r.ahmed@accenture.com}
\author{Sourabh Deshpande}
\affiliation{%
  \institution{Center for Advanced AI, Accenture}
  \city{Mountain View}
  \state{California}
  \country{USA}
}
\email{sourabh.a.deshpande@accenture.com}


\begin{abstract}
  Multi-agent frameworks (MAFs) promise to simplify LLM-driven software development, yet no principled metric captures how well AI coding assistants can generate correct, framework-specific code. We introduce \textit{AI-assistability} ($\mathcal{AI}$), a composite metric that quantifies a framework's amenability to AI-assisted development by combining structural alignment ($\bar{\sigma}$) with functional correctness (pass@1). To evaluate this metric in a controlled setting, we design DDL2PropBank, a novel benchmark task that maps relational database schemas to PropBank semantic rolesets, and implement identical agent logic across ten frameworks using the Agent-as-a-Tool pattern. Our results challenge the intuition that declarative framework design guarantees AI-assistability: Agno, with a single canonical pattern and convention-aligned API, achieves the highest $\mathcal{AI}$ score (0.55), while DSPy---the most declarative framework by design---scores lowest (0.07), as its novel abstractions are insufficiently represented in AI training data. We find that convention alignment, not declarative design alone, is the primary driver of AI-assistability ($r = 0.576$ between $\bar{\sigma}$ and pass@1).
  \footnote{All artifacts---DDL2PropBank, PropBank MCP server, and all implementations---are available at \url{https://github.com/ahmeshaf/ddl2propbank}. See \S\ref{appendix:artifact_summary} for the full list.}
\end{abstract}

\ccsdesc[500]{Computing methodologies~Multi-agent systems}
\ccsdesc[300]{Computing methodologies~Lexical semantics}
\ccsdesc[300]{Information systems~Relational database query languages}

\keywords{AI-assistability, multi-agent frameworks, developer experience, PropBank, schema mapping, LLM-as-Judge, Model Context Protocol, code generation}


\maketitle
\pagestyle{plain}   


\section{Introduction}
\label{sec:introduction}


In an LLM-based multi-agent system, ``multiple autonomous agents collaboratively engage in planning, discussions, and decision-making'' \cite{guo2024llmmultiagent}, coordinated through an orchestration layer that manages tool use and execution flow. Popular multi-agent frameworks (MAFs) such as LangChain, LangGraph, AutoGen, CrewAI, and others \cite{derouiche2025agentic} promise to streamline the development of these systems by providing reusable infrastructure for agent coordination and tool integration. However, as we demonstrate in \S\ref{sec:static_analysis}, these frameworks primarily supply orchestration primitives: users are still expected to write substantial code to define agent roles, compose workflows, handle control flow, and adapt abstractions to task-specific requirements. As the market for MAFs grows\footnote{The multi-agent system market is valued at \$10.6B in 2025, projected to exceed \$375B by 2034 \cite{marketus2025mas}.}, it raises a fundamental question: which frameworks has the least \textbf{code complexity} of assembling agentic workflows.

Modern AI coding assistants decouple MAFs' perceived usability from its code complexity by rapidly generating framework-specific scaffolding from documentation and examples \cite{liang2023largescalesurveyusabilityai}. This aligns with the ethos of \textit{vibe coding}\footnote{\url{https://en.wikipedia.org/wiki/Vibe_coding}}, in which implementation is increasingly delegated to AI assistants. While this accelerates adoption, it can also obscure substantive differences between frameworks: plausible-looking code may mask brittle abstractions or implicit control logic, and failures often manifest as hallucinated correctness or silent errors. Consequently, a framework's practical usability depends not only on its human-facing APIs, but also on \textit{the extent to which its abstractions and documentation enable AI coding assistants to generate correct, executable code}—a property we refer to as a framework's \textbf{AI-assistability}.

Evaluating these two aspects of developer experience---code complexity and AI-assistability---is a nontrivial challenge \cite{winters2020software,ziegler2024copilot}. Because LLMs drive code generation, planning, and control flow, conventional evaluations risk conflating framework quality with model memorization or benchmark familiarity. We therefore posit that evaluating developer experience requires a controlled task that is complex enough to stress-test AI-assisted code generation, yet uniform enough for fair comparison. Crucially, such a task must be novel with respect to LLM training data, avoiding settings where agents can trivially retrieve known solutions or exploit patterns from common benchmarks, thereby isolating framework-induced differences from artifacts of prior exposure.

To this end, we introduce \textbf{DDL2PropBank}, an agentic task of mapping database schemas to PropBank rolesets \cite{palmer2005propbank} corresponding to the events implicitly encoded by the database. Relational benchmarks like RelBench \cite{robinson2024relbench} formulate tasks as temporal next-event prediction, yet their schemas encode event semantics only implicitly; prior approaches such as RelGNN \cite{chen2025relgnncompositemessagepassing} recover event structure through representation learning over instance data rather than schema-level analysis. DDL2PropBank addresses this gap by requiring agents to derive explicit event representations directly from schema structure, constraints, and inter-table relations. While this task has potential uses, we treat it primarily as a novel, controlled, and sufficiently complex \textit{benchmarking vehicle} to compare MAFs.

\begin{figure}[t!]
    \centering
    \includegraphics[width=0.75\linewidth]{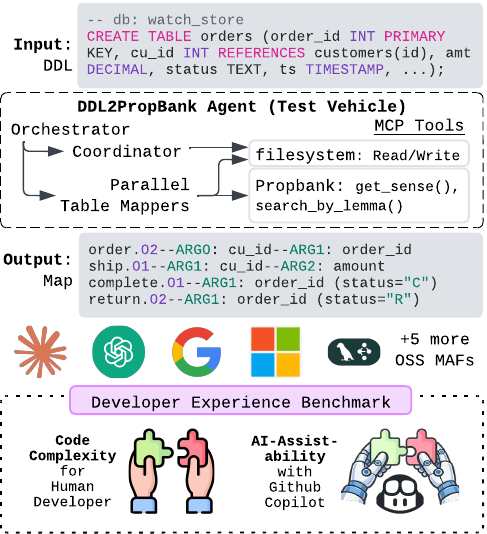}
    \caption{\textbf{Top:} The Agent-as-a-Tool architecture—an Orchestrator invokes a Coordinator and parallel Table Mapper agents, all accessing shared MCP servers (filesystem and PropBank). \textbf{Middle:} We implement identical logic across 10 MAFs including Claude SDK (Anthropic), Agents SDK (OpenAI), and 8 other open-source frameworks. \textbf{Bottom:} Dual-dimensional developer experience benchmarking.}
    \label{fig:overview}
\end{figure}

We adopt \textbf{Agent-as-a-Tool} design pattern, as illustrated in Figure~\ref{fig:overview}, where specialized agents are registered as callable tools by an orchestrator \cite{schick2023toolformer,li_survey_2024,li-2025-review}. The orchestrator first invokes a \textit{Coordinator} to track the progress of the mapping, then dispatches \textit{Table Mapper} agents parallelly to do semantic mapping for each table simultaneously. All agents interact with shared Model Context Protocol (MCP; \citet{anthropic2024mcp}) servers for filesystem access and PropBank queries. This architecture is intentionally generic and representative of real-world agentic usecases, and we expect it to be supported by any mature MAF.

Using human-authored reference implementations that fully specify MCP servers, prompts, and nested agent definitions, we evaluate both code complexity and AI-assistability. Code complexity is measured via static analysis across frameworks, using lines of code and cyclomatic complexity \cite{mccabe1976complexity} as proxies for the cognitive burden of assembling agentic workflows. To assess AI-assistability, we keep all functional components fixed and task \added{three AI coding assistants (GitHub Copilot, Claude Code, and Cursor)}\rc{369E} with reimplementing the same agent in each framework using only public APIs and documentation. We then evaluate whether the generated implementations (i) align structurally with the idiomatic human reference and (ii) execute end-to-end to produce \added{functionally valid PropBank mappings (correct tool grounding and well-formed output, which we term \emph{functional validity} rather than semantic optimality)}\rc{369E} on a test database.


Our results show that framework design choices lead to substantial variation in developer effort. Static analysis of the reference implementations reveals three complexity tiers: \textbf{\PydanticAI{}} and \textbf{\Agno{}} require the least code and control-flow logic; a middle tier includes \Smolagents{}, \LangChain{}, \OpenAIAgents{}, \GoogleADK{}, and \ClaudeSDK{}; and \AgentScope{} and \DSPy{} impose the highest overhead. For AI-assistability, structural alignment reliably predicts runtime robustness for frameworks with a single canonical pattern—\added{\Agno{} leads with the highest structural alignment and 72\% pass@1}\rc{369A,E}—but underestimates correctness for flexible, multi-pattern frameworks, \added{where non-idiomatic code can still execute (e.g.\ \PydanticAI{} reaches 61\% pass@1 at below-median alignment)}\rc{369A,E}. Overall, \textbf{\Agno{}} emerges as the strongest performer \added{($\mathcal{AI} = 0.55$)}, combining low code complexity with high structural alignment and robust functional correctness, making it particularly well-suited for idiomatic AI-assisted development.

Together, we make the following contributions:
\begin{itemize}[leftmargin=*, nosep]
    \item A first-of-its-kind benchmark, DDL2PropBank, for evaluating MAFs' developer experience.
    \item A dual-dimensional evaluation methodology assessing code complexity (via static analysis) and AI-assistability (via AI-generated code), offering actionable insights for framework developers and practical guidance for the NLP community.
    \item A novel event semantic analysis tool which makes use of the PropBank MCP server\footnote{Also introduced in this work in \S\ref{appendix:propbank_mcp}.} to facilitate semantic annotation of database schemas.
\end{itemize}

\section{DDL2PropBank Agent}
\label{sec:task}

\subsection{Task Definition}
\label{sec:task_definition}

Given a database schema in Data Definition Language (DDL), DDL2-PropBank is the semantic mapping of the tables in the schema to PropBank rolesets (See \S\ref{appendix:propbank} for a background on PropBank). Formally, the task is defined as follows:

\paragraph{Input.} Given a DDL schema $\mathcal{S} = \{T_1, T_2, \ldots, T_n\}$ where each table $T_i$ consists of columns $\{c_1, c_2, \ldots, c_m\}$ with types and constraints (primary keys, foreign keys).

\paragraph{Output.} For each table $T_i$, a set of mappings $\mathcal{M}_i = \{(r_j, A_j, \kappa_j)\}$ where:
\begin{itemize}[leftmargin=*, nosep]
    \item $r_j:$ PropBank roleset sense\_id (e.g., \texttt{order.02})
    \item $A_j: \{\texttt{ARG0}, \texttt{ARG1}, \ldots\} \rightarrow \{c_1, c_2, \ldots, c_m\}$ maps semantic arguments of $r_j$\ to columns
    \item $\kappa_j \in [0,1]$ is a confidence score reflecting the semantic fit of $(r_j, A_j)$
\end{itemize}

\paragraph{Example.} A single table can encode multiple event semantics, yielding several roleset mappings. As illustrated in Figure~\ref{fig:overview}, an \tbl{Orders} table with columns \col{cu\_id} (customer), \col{order\_id}, \col{amt} (amo-unt), \col{status}, and \col{ts} (order timestamp) maps to four rolesets---\texttt{order.02}, \texttt{ship.01}, \texttt{complete.01}, and \texttt{return.02}---each capturing a distinct event in the order lifecycle. For instance, \texttt{order.02} (``request to be delivered'') grounds ARG0 (orderer) $\rightarrow$ \col{cu\_id} and ARG1 (thing ordered) $\rightarrow$ \col{order\_id}, while \texttt{ship.01} grounds ARG1 $\rightarrow$ \col{cu\_id} and ARG2 $\rightarrow$ \col{amt}. Both rolesets also map ARGM-TMP (temporal) $\rightarrow$ \col{ts}. Arguments without a matching column are left unmapped, and each mapping receives a confidence score $\kappa_j$ reflecting its semantic fit.

\paragraph{\added{Novelty and Memorization.}} \added{To our knowledge DDL2PropBank is novel: no prior work maps relational schemas to PropBank rolesets, and PropBank access is exposed through a custom-built MCP server (\S\ref{appendix:propbank_mcp}) rather than any pre-existing API. The combination---Agent-as-a-Tool orchestration, shared MCP servers, and schema-level semantic annotation---has no precedent in public training data, mitigating memorization confounds: assistants cannot retrieve a known reference solution, so generated implementations reflect each framework's documentation and abstractions rather than recall of the task.}\rc{369E}



\subsection{Reference Agent-as-a-Tool Architecture}
\label{sec:architecture}

We implement DDL2PropBank using the Agent-as-a-Tool pattern, where specialized ReAct-style\footnote{ReAct, a portmanteau of \textbf{Re}asoning and \textbf{Act}ing, is an approach introduced by \citet{yao2023react} that enables language models to ``generate both reasoning traces and task-specific actions in an interleaved manner.''} agents are registered as callable tools within a orchestrator. The architecture comprises three agent types operating at different granularities.\footnote{See \S\ref{appendix:agent_design} for algorithmic details of the agent design, including pseudocode and prompts.}

\subsubsection{Agent Descriptions}
\label{sec:agent_descriptions}

\paragraph{Orchestrator.} The top-level agent manages the complete workflow for a single database. It invokes the Coordinator to identify pending work, then dispatches Table Mapper agents in parallel. This design enables incremental progress: if execution fails mid-way, previously completed mappings persist and the Coordinator will skip them on retry.

\paragraph{Coordinator.} The Coordinator examines the output folder to determine which tables have valid mappings and which require processing. By checking filesystem state, it enables idempotent execution---re-running the pipeline only processes missing or invalid mappings.

\paragraph{Table Mapper.} Each Table Mapper processes a single table independently. It extracts column information and foreign key context from the DDL, invokes an LLM-tool to generate candidate action verbs from table context (e.g., \tbl{Orders} $\rightarrow$ \texttt{["order", "purchase"]}), queries PropBank via MCP for matching rolesets, then grounds semantic arguments to columns. For each candidate mapping, the agent estimates a confidence score $\kappa_j \in [0,1]$ based on semantic fit quality. The complete mapping set $\mathcal{M}_T = \{(r_j, A_j, \kappa_j)\}$ is persisted immediately upon completion.

\subsubsection{MCP Integration}
\label{sec:mcp_integration}

The Model Context Protocol (MCP) is an open standard for connecting AI assistants to external data sources and tools via standardized message passing. By exposing PropBank and filesystem operations as MCP servers, we decouple agent logic from tool infrastructure, enabling all 10 frameworks to access identical tool capabilities. This architectural choice is critical: agents query the same PropBank server, read the same schemas, and write outputs to the same filesystem, ensuring that observed differences in mapping quality arise solely from framework-specific choices.

\paragraph{PropBank MCP Server (StreamableHTTP).} Provides programmatic access to PropBank's curated rolesets, avoiding reliance on LLM memorization of semantic frame definitions\footnote{See \S\ref{appendix:propbank_mcp} for details on the motivation and implementations of the PropBank MCP server.}:
\begin{itemize}[leftmargin=*, nosep]
    \item \texttt{search\_by\_lemma(word, max\_res)}: Query PropBank for all rolesets matching a verb base form or morphological variant
    \item \texttt{search\_by\_sense\_id(sense\_id, include\-\_examples)}: Retrieve full frame def for the roleset, including roles and annotated examples
\end{itemize}

\paragraph{Filesystem MCP Server (StdIO).} Manages agent state and output persistence, enabling incremental progress and fault tolerance:
\begin{itemize}[leftmargin=*, nosep]
    \item \texttt{list\_directory()}: Enumerate files in the output folder to check mapping progress.
    \item \texttt{read\_file()}: Load previously computed mappings for validation or reuse
    \item \texttt{write\_file()}: Persist new mappings immediately upon completion
\end{itemize}

\paragraph{Get Action Verbs function tool.} An LLM-tool that generates candidate verbs from table context via a single LLM call. Given a table's schema. the tool prompts the LLM to produce lemmas capturing the table's potential event semantics---e.g., \tbl{Orders} may give \texttt{["order", "ship", "cancel", "return"]}. These lemmas seed PropBank queries, enabling exploration of semantic frames (see Fig. \ref{fig:getactionverbs_prompt} in \S\ref{appendix:system_prompts}).


\begin{figure*}[t]
\centering
\resizebox{0.9\textwidth}{!}{%
\begin{minipage}{\textwidth}
\begin{minipage}[t]{0.48\textwidth}
\begin{algorithm}[H]
\caption{Orchestrator}
\label{alg:orchestrator}
\begin{algorithmic}[1]
\Require DDL schema $\mathcal{S}$, output folder $\mathcal{F}$
\Ensure PropBank mappings for all tables

\State $\mathcal{T}_{todo} \gets$ \Call{Coordinator}{$\mathcal{S}$, $\mathcal{F}$} \atoolbadge \label{step:coordinator}
\ParFor{$T \in \mathcal{T}_{todo}$} \llmbadge
    \State \Call{TableMapper}{$T$, $\mathcal{S}$, $\mathcal{F}$} \atoolbadge
\EndParFor
\State \textbf{go to} line~\ref{step:coordinator} \llmbadge \label{step:goto}
\end{algorithmic}
\end{algorithm}

\vspace{-2.2em}

\begin{algorithm}[H]
\caption{Coordinator}
\label{alg:coordinator}
\begin{algorithmic}[1]
\Require DDL schema $\mathcal{S}$, output folder $\mathcal{F}$
\Ensure Tables requiring mapping ($\mathcal{T}_{todo}$)
\State $\mathcal{T}_{all} \gets$ \textbf{parse} table names from $\mathcal{S}$ \llmbadge
\For{$T \in \mathcal{T}_{all}$} \llmbadge
    \State $\mathcal{M}_T \gets$ \Call{ReadFile}{$\mathcal{F}/T.\text{json}$}\toolbadge
    \If{\textbf{not} valid $\mathcal{M}_T$} \llmbadge
        \State $\mathcal{T}_{todo} \gets \mathcal{T}_{todo} \cup \{T\}$
    \EndIf
\EndFor
\State \Return $\mathcal{T}_{todo}$
\end{algorithmic}
\end{algorithm}
\end{minipage}%
\hfill%
\begin{minipage}[t]{0.48\textwidth}

\begin{algorithm}[H]
\caption{TableMapper}
\label{alg:tablemapper}
\begin{algorithmic}[1]
\Require Table $T$, schema $\mathcal{S}$, output folder $\mathcal{F}$
\Ensure Mapping $\mathcal{M}$ persisted to filesystem
\vspace{1mm}
\State $context \gets$ \textbf{analyze} $\mathcal{S}$ w.r.t $T$ \llmbadge
\vspace{1mm}

\label{alg:line:getverbs}
\State $lemmas \gets$ \Call{GetActionVerbs}{$T$, $ctx$} \llmtoolbadge 
\vspace{1mm}
\For{$lemma \in lemmas$} \llmbadge
\vspace{1mm}
    \State $rolesets \gets$ \Call{SearchByLemma}{$lemma$} \toolbadge
    \vspace{1mm}
    \State $\mathcal{R} \gets$ \textbf{identify} relevant $rolesets$ \llmbadge
    \vspace{1mm}
    \For{$r \in \mathcal{R}$} \llmbadge 
        \vspace{1mm}
        \State $args \gets$ \Call{SearchBySenseId}{$r$} \hphantom{xxxxx} \toolbadge
        \vspace{2mm}
        \State $\mathcal{M} \gets \mathcal{M} \cup \{(r,$ \textbf{map} $args$ to $context$, \textbf{estimate} $\kappa)\}$ \hphantom{xxx} \llmbadge
    \EndFor
\vspace{1mm}
\EndFor
\vspace{1mm}
\State $\mathcal{M}_{T} \gets$ \textbf{select} the best mappings from $\mathcal{M}$ that are \textbf{sorted by} $\kappa$ \hphantom{xxx} \llmbadge
\vspace{1mm}
\State \Call{WriteFile}{$\mathcal{F}/T.\text{json}$, $\mathcal{M}_T$} \toolbadge
\end{algorithmic}
\end{algorithm}
\end{minipage}%
\end{minipage}%
}
\caption{DDL2PropBank Agent algorithms for Agent-as-a-Tool pattern using only the agentic primitives defined in \S\ref{appendix:step_types}.
 }
\label{fig:algorithms}
\end{figure*}

\subsection{Pseudocode and Agentic Primitives}
\label{appendix:agent_algorithms}
\label{appendix:step_types}

Figure~\ref{fig:algorithms} presents the pseudocode for the three DDL2PropBank agents. The Orchestrator (Algorithm~\ref{alg:orchestrator}) serves as the top-level coordinator, invoking Coordinator and TableMapper as callable tools. The Coordinator (Algorithm~\ref{alg:coordinator}) analyzes existing mappings to determine which tables require processing. The TableMapper (Algorithm~\ref{alg:tablemapper}) handles per-table PropBank mapping, iterating over candidate verbs, searching for matching rolesets, and persisting results. We identify four \emph{agentic primitives}---atomic building blocks of the agent algorithms, ordered here by increasing autonomy:

\paragraph{\fcolorbox{OliveGreen}{green!15}{\textcolor{OliveGreen}{tool}}} A direct call to an MCP server tool such as \texttt{ListDirectory}, \texttt{ReadFile}, \texttt{WriteFile}, \texttt{SearchByLemma}, or \texttt{SearchBySenseId}. These are deterministic operations with no LLM reasoning involved.

\paragraph{\fcolorbox{OliveGreen}{green!15}{\textcolor{OliveGreen}{LLM-tool}}} A tool encapsulating a \emph{single} LLM inference. For example, \texttt{GetActionVerbs} infers action verbs from table context. Callers treat these as tools, but they internally invoke one LLM call.

\paragraph{\fcolorbox{TealBlue}{TealBlue!15}{\textcolor{TealBlue}{Agent-as-a-tool}}} A sub-agent registered as a callable tool. The Orchestrator invokes \texttt{Coordinator} and \texttt{TableMapper} as tools, enabling hierarchical composition and parallel execution via concurrent tool calls.

\paragraph{\fcolorbox{violet}{violet!15}{\textcolor{violet}{ReAct}}} LLM reasoning steps encoded directly in the system prompt, without external code logic. The agent iteratively analyzes context, makes decisions, and synthesizes information---relying solely on the driving LLM's reasoning capabilities. Critically, ReAct governs execution topology, shaping the agentic workflow by determining whether tool calls are issued in parallel or sequentially.

These primitives define a framework-agnostic vocabulary for agentic workflows, however, multi-agent frameworks may differ substantially in how they realize each one---differences that directly impact code complexity and developer effort (\S\ref{sec:ground_truth}).


\subsection{Ground-truth Implementations}
\label{sec:ground_truth}

We implement the DDL2PropBank agent architecture across 10 multi-agent frameworks, creating human-authored reference implementations that serve as ground truth for our AI-assistability evaluation (\S\ref{sec:copilot}). Each implementation realizes identical agent logic---the Orchestrator, Coordinator, and Table Mapper hierarchy described in \S\ref{sec:architecture}---using only framework-native abstractions. System prompts are standardized across all frameworks (App~\ref{appendix:system_prompts}), ensuring that observed differences in code complexity and runtime behavior arise solely from framework design choices rather than algorithmic variation.

\subsubsection{Framework Selection}
\label{sec:framework_selection}

We include all three major LLM provider frameworks---\ClaudeSDK{}, \OpenAIAgents{}, and \GoogleADK{}---as first-party agentic solutions from the primary model providers. For the remaining seven community-maintained frameworks (\PydanticAI{}, \Agno{}, \DSPy{} \cite{khattab2024dspy}, \LangChain{}, \Microsoft{} Agent Framework, \Smolagents{} \cite{smolagents}, \AgentScope{} \cite{gao2025agentscope}), we applied eight selection criteria spanning four dimensions:
\begin{itemize}[leftmargin=*, nosep]
    \item \textit{Reproducibility}: open-source licensing (C1) and programmatic invocation for batch processing (C7).
    \item \textit{Tool integration}: support for both MCP servers and direct function calling (C2), with concurrent tool invocations for parallel processing (C3).
    \item \textit{Operational}: multi-provider model support (C5), single API key operation (C6), and programmatic access to token usage metrics (C4).
    \item \textit{AI-assistability}: comprehensive documentation (C8), with \texttt{llms.txt} availability prioritized for AI-assisted code generation.
\end{itemize}
See Table~\ref{tab:framework-criteria} in \S\ref{appendix:framework_comparison} for per-framework compliance with each criterion.

\subsubsection{Static Code Analysis}
\label{sec:static_analysis}

To quantify the structural complexity of assembling agentic workflows, we analyze each framework's \texttt{database\_mapper.py} implementation using standard static analysis metrics (Table~\ref{tab:code_complexity}). Logical lines of code (LLOC) measures implementation size excluding comments and blank lines; values in parentheses indicate lines added beyond the 31-line starter template. Cyclomatic complexity (CCN) captures control flow branching---higher values indicate more complex decision logic. Framework-specific imports reflect API surface area and coupling to framework internals.

\paragraph{Low-Complexity Winners.}
\textbf{Pydantic AI} and \textbf{Agno} emerge as co-leaders in code complexity. Pydantic AI achieves lower LLOC (52 vs.\ 54) and CCN (3 vs.\ 4), while Agno requires fewer imports (3 vs.\ 4).
Their advantage stems from a shared set of design choices that systematically reduce boilerplate. In particular, both frameworks support \emph{unified tool declarations} that allow MCP servers and Python functions to coexist without wrapper code, \emph{decorator-free function registration} that accepts plain Python functions, \emph{model-string--based provider inference} that avoids provider-specific client classes, and \emph{native MCP abstractions} that manage connection lifecycles and expose token usage without custom instrumentation. Collectively, these patterns minimize scaffolding code while preserving expressiveness, resulting in more concise and maintainable agentic solution.

\begin{table}[t!]
\caption{Code complexity metrics for agent implementations. LLOC = logical lines of code; CCN = cyclomatic complexity; \#Fns = number of functions; Imports = framework-specific imports. Bold indicates best (lowest).}
\centering
\small
\begin{tabular}{@{}llccc@{}}
\toprule
\textbf{Framework} & \textbf{LLOC} & \textbf{CCN} & \textbf{\#Fns} & \textbf{Imports} \\
\midrule
\multicolumn{5}{@{}l}{\textit{LLM Providers' Frameworks}} \\
\midrule
Google ADK & 68 {\scriptsize(+37)} & 12 & 4 & 8 \\
OpenAI Agents & 70 {\scriptsize(+39)} & 14 & 8 & 5 \\
Claude SDK & 75 {\scriptsize(+44)} & 10 & 5 & 7 \\
\midrule
\multicolumn{5}{@{}l}{\textit{Independent Frameworks}} \\
\midrule
Pydantic AI & \textbf{52} {\scriptsize(+21)} & \textbf{3} & \textbf{3} & 4 \\
Agno & 54 {\scriptsize(+23)} & 4 & \textbf{3} & \textbf{3} \\
Smolagents & 66 {\scriptsize(+35)} & 15 & 5 & 6 \\
\LangChain{} & 70 {\scriptsize(+39)} & 11 & 4 & 8 \\
Microsoft & 73 {\scriptsize(+42)} & 16 & 4 & 4 \\
AgentScope & 82 {\scriptsize(+51)} & 10 & 5 & 10 \\
DSPy & 88 {\scriptsize(+57)} & 20 & 6 & 9 \\
\midrule
\textit{Starter code} & 31 & - & 3 & - \\
\bottomrule
\end{tabular}
\label{tab:code_complexity}
\end{table}

\paragraph{Mid-Complexity Frameworks.}
The six remaining frameworks---all three vendor SDKs (\GoogleADK{}, \OpenAIAgents{}, \ClaudeSDK{}) and three independent projects (\Smolagents{}, \LangChain{}, \Microsoft{})---cluster in the LLOC 66--75 range. The vendor SDKs share comparable implementation size but differ in control-flow complexity: \GoogleADK{} and \ClaudeSDK{} exhibit moderate CCN (10--12), while \OpenAIAgents{} requires more branching (CCN 14) due to explicit tool-result handling. Among the independent frameworks, \Smolagents{} and \Microsoft{} exhibit the highest CCN in this tier (15--16), reflecting more verbose control flow for agent orchestration and tool registration.

\paragraph{High-Complexity Frameworks.}
\textbf{DSPy} exhibits the highest code complexity---LLOC of 88 and CCN of 20---driven primarily by the absence of native MCP support. As a result, developers must integrate external MCP libraries, manage server lifecycles manually and adapt MCP tools into framework-specific abstractions. \textbf{AgentScope} exhibits lower control-flow complexity but incurs substantial overhead from extensive framework imports: 10 imports across seven submodules, compared to 3--4 for lower-complexity frameworks. Reliance on internal abstractions for formatters, memory classes, tool response objects, and provider-specific model classes increases cognitive load and tightly couples implementations to framework internals.

Across frameworks, we observe a 1.7$\times$ variation in LLOC and a 3.3$\times$ variation in import count for implementing identical agent behavior, underscoring that framework design choices—particularly native MCP integration and unified tool registration—have a substantial impact on developer effort. For detailed analysis of implementation patterns across all frameworks see \S\ref{appendix:code_complexity_analysis}.


\section{Framework's AI-Assistability Evaluation}
\label{sec:copilot}

AI coding assistants are transforming software development. A controlled experiment by \citet{ziegler2024copilot} found that developers using GitHub Copilot completed tasks 55.8\% faster than those without access, with junior developers showing the largest gains. As tools like GitHub Copilot, Claude Code, and Cursor become ubiquitous, a framework's raw code complexity---the LLOC and CCN metrics from \S\ref{sec:static_analysis}---becomes less decisive. If an AI assistant can reliably generate framework specific boilerplate, developers need not manually navigate verbose APIs or complex control flow. The more pressing question shifts from \textit{how much code must I write?} to \textit{how well can AI assistants write it for me?}

This reframing motivates our second evaluation dimension: \textbf{AI-assistability}---the extent to which large language models can autonomously generate correct, framework-specific code from documentation alone. We hypothesize that frameworks differ substantially in AI-assistability due to variations in (i)~documentation quality (comprehensive API references vs.\ scattered tutorials), (ii)~API consistency (predictable patterns vs.\ special cases), (iii)~abstraction level (declarative vs.\ imperative control flow), and (iv)~error message clarity (actionable diagnostics vs.\ opaque traces). To isolate these framework-intrinsic factors, we formalize the evaluation as follows.

\paragraph{Generation Context.} Let $\mathcal{G} = (a, m, \tau, P_\tau)$ be a generation context: AI coding assistant $a \in \mathcal{A}$, model tier $m \in \mathcal{M}$, task $\tau$, and its associated project artifacts $P_\tau$ (task specification, starter code, agent prompts, entry point). Note that $P_\tau$ can vary with the expertise level of the developer: a more experienced practitioner may provide richer task specifications, more detailed prompts, or better-structured starter code, all of which influence generation quality independently of the framework or assistant.
We vary $a$, $m$, and the framework~$f$; the only framework specific input is the documentation $D_f$. The assistant generates:
$$\hat{y}_f \sim a(D_f, P_\tau;\; m)$$

\paragraph{Validation.} Let $y_f^*$ denote the human-authored ground-truth implementation for framework $f$ (\S\ref{sec:ground_truth}). Each $\hat{y}_f$ is evaluated by two complementary functions:
\begin{itemize}[leftmargin=*, nosep]
    \item $\sigma(\hat{y}_f, y_f^*) \in [0, 24]$: \textit{structural alignment}---LLM-as-Judge score against $y_f^*$ (rubric in \S\ref{sec:llm_judge})
    \item $\phi(\hat{y}_f) \in \{0, 1\}$: \textit{functional correctness}---runtime pass/fail on a test database (\S\ref{sec:runtime_testing})
\end{itemize}

For a fixed configuration $\mathcal{G}$, we compute per-run metrics over $n$ independent runs:
$$\text{pass@1}(f \mid \mathcal{G}) = \frac{1}{n}\sum\nolimits_{i=1}^{n} \phi(\hat{y}_{f,i}), \quad \bar{\sigma}(f \mid \mathcal{G}) = \frac{1}{n}\sum\nolimits_{i=1}^{n} \sigma(\hat{y}_{f,i}, y_f^*)$$

To obtain a single score per framework, we marginalize over assistants and model tiers by averaging across all configurations:
\begin{align}
\text{pass@1}(f) &= \frac{1}{|\mathcal{A}||\mathcal{M}|}\sum\nolimits_{a,m} \text{pass@1}(f \mid \mathcal{G}) \\
\bar{\sigma}(f) &= \frac{1}{|\mathcal{A}||\mathcal{M}|}\sum\nolimits_{a,m} \bar{\sigma}(f \mid \mathcal{G})
\end{align}

\paragraph{AI-Assistability Score.} We define a framework's \textit{AI-assistability} as the product of its normalized structural alignment and functional correctness:
$$\mathcal{AI}(f) = \text{pass@1}(f) \;\times\; \frac{\bar{\sigma}(f)}{\sigma_{\max}}$$
where $\sigma_{\max} = 24$ is the maximum rubric score. This multiplicative formulation requires \textit{both} dimensions to be high: a framework with perfect structural alignment but zero pass rate scores $\mathcal{AI} = 0$, and vice versa. The score ranges from 0 (complete failure on either dimension) to 1 (perfect on both).

By varying $f$, $a$, and $m$, differences in $\mathcal{AI}$, pass@1, and $\bar{\sigma}$ isolate framework-intrinsic properties from assistant- and model-specific effects. Although we instantiate $\tau$ = DDL2PropBank in this work, the formalization is task-agnostic: substituting a different agentic task $\tau'$ with corresponding artifacts $P_{\tau'}$ and ground-truth implementations $y_{f}^{*\prime}$ yields a reusable evaluation protocol for any MAF comparison.



\subsection{Evaluation Protocol ($\mathcal{A}$, $D_f$)}
\label{sec:copilot_workflow}

We evaluate three AI coding assistants, $\mathcal{A} = \{\text{GitHub Copilot},$ $\text{Claude Code}, \text{Cursor}\}$, each configured with Claude models via Anthropic's API. All three operate in \textit{agent mode}, where they can read project files, browse documentation URLs, and iteratively edit code---capabilities beyond single-turn code completion. This agentic workflow mirrors realistic developer usage: the assistant explores the codebase, consults framework documentation, and produces a complete implementation over multiple tool-calling steps. The three assistants represent distinct integration paradigms: GitHub Copilot is IDE-embedded (VS Code), Claude Code is CLI-native, and Cursor is an AI-first IDE fork. By varying the assistant alongside the framework, we disentangle framework-intrinsic difficulty from assistant-specific strengths (\S\ref{sec:code_gen_config}).

Figure~\ref{fig:copilot_architecture} illustrates the evaluation protocol. For each framework $f$, a developer provides a standardized \textbf{query template} that supplies the framework-specific documentation $D_f$: the framework name, its documentation URL, and its GitHub repository URL. This query is the sole input that varies across frameworks. The assistant then reads the shared project context $P$---task specification (\texttt{CLAUDE.md}\footnote{\texttt{CLAUDE.md} is a convention for providing project-specific instructions to coding assistants. See \url{https://docs.anthropic.com/en/docs/claude-code}.}), agent prompts (\texttt{prompts.py}), and the starter code with TODOs---and generates the framework-specific implementation $\hat{y}_f$ in \texttt{database\_mapper.py}.

\begin{figure}[t]
\centering
\resizebox{0.84\columnwidth}{!}{
\begin{tikzpicture}[
    box/.style={rectangle, draw, rounded corners, font=\footnotesize, align=left, inner sep=3pt},
    container/.style={rectangle, draw, rounded corners, inner sep=4pt},
    arrow/.style={-{Stealth[length=2mm]}, thick},
]

\node[box, fill=orange!15, minimum width=7cm, minimum height=0.6cm, align=center] (assistant) {\textbf{AI Coding Assistant $a \in \mathcal{A}$}};

\node[box, above=0.4cm of assistant.north west, anchor=south west, fill=white, text width=3.2cm] (query) {
    {\footnotesize\textbf{Query Template ($D_f$)}}\\[1pt]
    {\scriptsize framework: \texttt{\{name\}}\\
    documentation page: \texttt{\{url\}}\\
    github page: \texttt{\{url\}}\\[-0.5mm]
    additional constraints: \texttt{[...]}}
};

\node[box, above=0.4cm of assistant.north east, anchor=south east, fill=white, text width=3.2cm] (cwd) {
    {\footnotesize\textbf{Project Directory ($P_\tau$)}}\\[2pt]
    \fcolorbox{blue!50}{blue!10}{\makebox[2.95cm][l]{\scriptsize\pencil~database\_mapper.py}}\\[2pt]
    \fcolorbox{gray}{gray!15}{\makebox[2.95cm][l]{\scriptsize\lock~read-only files}}
};


\node[box, below=0.35cm of assistant, fill=green!10, align=center, minimum width=7cm, minimum height=0.6cm] (output) {$\hat{y}_f$: database\_mapper.py \footnotesize(implemented)};


\draw[arrow] (query.south) -- (query.south|-assistant.north);
\draw[arrow] (cwd.south) -- (cwd.south|-assistant.north);
\draw[arrow] (assistant) -- (output);

\end{tikzpicture}
}
\caption{Evaluation protocol for AI-assistability. The query template provides framework-specific documentation $D_f$; the project directory provides task artifacts $P_\tau$. Each assistant $a \in \mathcal{A}$ generates $\hat{y}_f$ by implementing \texttt{database\_mapper.py} (\pencil), with all other files locked (\lock).}
\label{fig:copilot_architecture}
\end{figure}

The workflow enforces a clear separation: each assistant implements only the framework integration layer (\texttt{database\_mapper.py}), while agent behavior (system prompts), task specification, and tool infrastructure remain fixed across all frameworks and assistants. This isolation ensures that observed differences in pass@1 and $\bar{\sigma}$ stem from framework and assistant characteristics rather than task variation.

\subsection{Project Structure ($P_\tau$)}
\label{sec:project_structure}

Each framework evaluation uses an identical project scaffold (Figure~\ref{fig:project_structure}). The assistant must understand task requirements from \texttt{CLAUDE.md}, reuse agent prompts from \texttt{prompts.py}, and produce code that integrates with the fixed \texttt{main.py} entry point.

\begin{figure}[h]
\centering
\begin{minipage}{\columnwidth}
\raggedright
\small
\dirtree{%
.1 src\_\{framework\}/.
.2 CLAUDE.md \lock{} \textrm{\scriptsize\textcolor{gray}{-- task spec, architecture}}.
.2 prompts.py \lock{} \textrm{\scriptsize\textcolor{gray}{-- agent prompts (641 LOC)}}.
.2 main.py \lock{} \textrm{\scriptsize\textcolor{gray}{-- entry point}}.
.2 examples/.
.3 rel-avito.sql \lock{} \textrm{\scriptsize\textcolor{gray}{-- 8-table test schema}}.
.2 \textbf{database\_mapper.py} \pencil{} \textrm{\scriptsize\textbf{-- 31-line stub w/ TODOs}}.
}
\end{minipage}
\caption{Project structure for AI-assisted implementation. Files marked \lock\ are read-only context; the assistant implements only \texttt{database\_mapper.py} (\pencil).}
\label{fig:project_structure}
\end{figure}

The \texttt{database\_mapper.py} file is the sole implementation target---a 31-line stub containing function signatures, docstrings, and TODO comments that the assistant must complete using framework-native abstractions (see Listing~\ref{lst:skeleton} for the full skeleton). The stub specifies three components: \texttt{run\_coordinator()} for progress analysis, \texttt{run\_mapper()} for single-table PropBank mapping, and orchestration logic that coordinates these via LLM tool-calling. This decomposition directly reflects the three-agent hierarchy from \S\ref{sec:architecture}, with each stub function corresponding to an agent the assistant must realize using framework-native APIs.


\subsection{Experimental Configuration ($m$, $n$)}
\label{sec:code_gen_config}

We vary the model tier $m$ in the generation context $\mathcal{G}$ to test two capability levels.\footnote{The DDL2PropBank agents themselves use pre-specified model assignments detailed in \S\ref{appendix:model_config}.}

\paragraph{Model Tiers.} We test two Claude model tiers as the coding assistant backend: Claude \textbf{Haiku 4.5} (fast, cost-effective) and Claude \textbf{Sonnet 4.5} (balanced capability).
All three assistants are configured with the same Claude model for a given tier, isolating assistant-level differences from model capability. This comparison reveals whether framework complexity creates a capability threshold---some frameworks may require more capable models to implement correctly.

\paragraph{Replication Strategy.} To account for stochastic variation in LLM generation, we execute $n = 3$ independent runs per configuration, yielding $10 \text{ frameworks} \times 3 \text{ assistants} \times 2 \text{ models} \times 3 \text{ runs} = 180$ implementations total. Each assistant is allowed to run until reaching context window compaction---the point at which the assistant's conversation history is compressed to fit within its context limit---ensuring that generation is bounded by the assistant's effective memory rather than an artificial step cap. Each run provides the same inputs---starter code, task specification (\texttt{CLAUDE.md}), and framework documentation URLs---but generates agent code independently.


\subsection{Evaluation Metrics ($\sigma$, $\phi$)}
\label{sec:evaluation_methodology}

Generated implementations are evaluated along two independent dimensions: structural alignment $\sigma$ via automated LLM-based assessment and functional correctness $\phi$ via human-supervised runtime testing.

\subsubsection{LLM-as-Judge: Structural Alignment ($\sigma$)}
\label{sec:llm_judge}

We employ Claude Opus 4.5 as an automated code reviewer, comparing each generated implementation against human-authored references (described in \S\ref{sec:ground_truth}). This alignment-based evaluation measures whether AI assistants can autonomously recreate idiomatic code from documentation alone. The rationale is twofold: idiomatic code is (i) \textit{maintainable}---adhering to established patterns makes code easier to debug, extend, and review; and (ii) \textit{trustworthy}---our ground-truth solutions are validated through unit and end-to-end testing, so AI-generated code that follows the same patterns inherits this robustness. High structural alignment thus serves as a proxy for robustness.

\added{Our LLM-judge-over-a-structured-rubric approach follows established evaluation practice, reviewed in \S\ref{sec:related}; it suits our setting because the property of interest is \emph{structural}---tool-registration patterns, MCP lifecycle, idiomatic API usage---rather than the token overlap that reference-based metrics capture. We anchor $\sigma$ to a single canonical reference per framework by design: the metric measures convergence toward canonical API usage rather than whether the assistant found \emph{any} valid pattern, so functionally correct but non-canonical code is flagged as divergent, e.g.\ Pydantic AI's high pass@1 at below-median $\bar{\sigma}$ (\S\ref{sec:outlier_analysis}).}\rc{369A,E}

\begin{table}[t]
\caption{LLM-as-Judge rubric for structural code alignment. MCP integration and API usage use weighted scoring (0--4) as they strongly predict runtime validity.}
    \centering
\resizebox{0.9\columnwidth}{!}{

\small
\begin{tabular}{@{}lcp{4cm}@{}}
\toprule
\textbf{Component} & \textbf{Scale} & \textbf{Evaluation Focus} \\
\midrule
\multicolumn{3}{@{}l}{\textit{Agent Architecture (0--3 each)}} \\
Coordinator & 0--3 & Inner agent with filesystem-readonly access \\
Mapper & 0--3 & Inner agent with full toolset \\
get\_verbs & 0--3 & LLM-tool in mapper's toolset \\
\midrule
\multicolumn{3}{@{}l}{\textit{MCP Integration (0--4 each, weighted)}} \\
filesystem\_mcp & 0--4 & StdIO server, allowed\_dirs \\
propbank\_mcp & 0--4 & HTTP server, correct endpoint \\
\midrule
\multicolumn{3}{@{}l}{\textit{Framework Integration}} \\
matching\_imports & 0--3 & Imports match ground truth \\
api\_usage & 0--4 & Idiomatic API patterns \\
\midrule
\textbf{Total} & \textbf{0--24} & Weighted sum \\
\bottomrule
\end{tabular}
}
\label{tab:judge_rubric}
\end{table}

The judge evaluates 7 architectural components critical to the Agent-as-a-Tool pattern as depicted in Table~\ref{tab:judge_rubric}. MCP integration and API usage receive higher weights (0--4) because they most strongly predict runtime validity: MCP configuration errors cause 73\% of runtime failures, while API misuse accounts for another 18\%. To reduce stochastic variance, each implementation receives 10 independent evaluations; we report median scores per component, then sum to compute totals.

\subsubsection{Runtime Validity Testing ($\phi$)}
\label{sec:runtime_testing}

A human executes each run on the \textbf{rel-avito} database from RelBench \citep{robinson2024relbench}---an 8-table classified advertisements schema that exercises all DDL2Prop\-Bank components (details in Appendix~\ref{appendix:test_database}). An implementation is marked \textbf{valid} if it (i) completes without exceptions, (ii) \added{correctly invokes the PropBank MCP tools (\texttt{search\_by\_lemma}, \texttt{search\_by\_\-sense\_id}) to ground mappings,} and (iii) produces correctly-structured JSON mappings for all 8 tables; otherwise it is \textbf{invalid}. \added{We therefore read $\phi$ as \emph{functional validity}---correct tool use, well-formed output, and end-to-end pipeline completion---rather than semantic optimality of the chosen rolesets, which would require gold annotations that do not exist for this novel task.}\rc{369E} This binary signal provides label for functional correctness, independent of structural assessment.\added{\footnote{Each run uses a generous framework-specific timeout scaled to the framework's ground-truth runtime rather than a single fixed budget, so a correct-but-slower framework is not penalized for speed; ground-truth runtimes (Table~\ref{tab:cost_effectiveness}) were measured without a timeout.}}\rc{369E} Frameworks achieving both high validity rates and high judge scores demonstrate robust AI-assistability---the ideal production sweet spot.



\section{Results}
\label{sec:results}

\subsection{Overall AI-Assistability}
\label{sec:overall_results}

GitHub Copilot produces valid DDL2PropBank implementations 66.7\% of the time (40/60 runs\added{; the GitHub Copilot subset of the full 180-run set, reported here as a single-assistant baseline---per-framework aggregates over all three assistants appear in Figure~\ref{fig:framework_scatter}}\rc{369E}) across all framework and model configurations. This two-thirds success rate establishes a practical baseline for AI-assisted multi-agent development: developers can expect functional code in the majority of attempts, though manual intervention remains necessary for roughly one-third of generations.

\textbf{Structural alignment strongly predicts runtime success} (Figure~\ref{fig:judge_stacked_bars}, Appendix~\ref{appendix:score_distribution}). Implementations scoring 18+ achieve 89\% pass@1\footnote{pass@k measures the probability that at least one of k generated code samples passes all tests \cite{chen2021evaluating}; we report pass@1, the strictest single-attempt metric.} (24/27), while those scoring below 12 achieve only 29\% (2/7). The Pearson correlation of $r = 0.576$ between judge score and pass@1 confirms that the LLM-as-Judge methodology captures meaningful code quality dimensions---high-scoring runs are substantially more likely to execute correctly. \added{We read this as a moderate relationship: $\bar{\sigma}$ captures a significant but partial dimension of functional correctness, and the remaining unexplained variance reflects runtime integration issues invisible to static structural analysis---MCP connection lifecycle, async compatibility, and session management---the failure modes that dominate Google ADK's case (\S\ref{sec:framework_landscape}).}\rc{369D} This correlation is strongest for frameworks with canonical implementation patterns: for Agno, Claude SDK, and OpenAI Agents, high structural alignment reliably implies correct tool registration and idiomatic API usage.


\subsection{Framework Landscape}
\label{sec:framework_landscape}

\begin{figure*}[t]
\centering
\begin{minipage}[c]{0.48\textwidth}
\centering
\includegraphics[width=\linewidth]{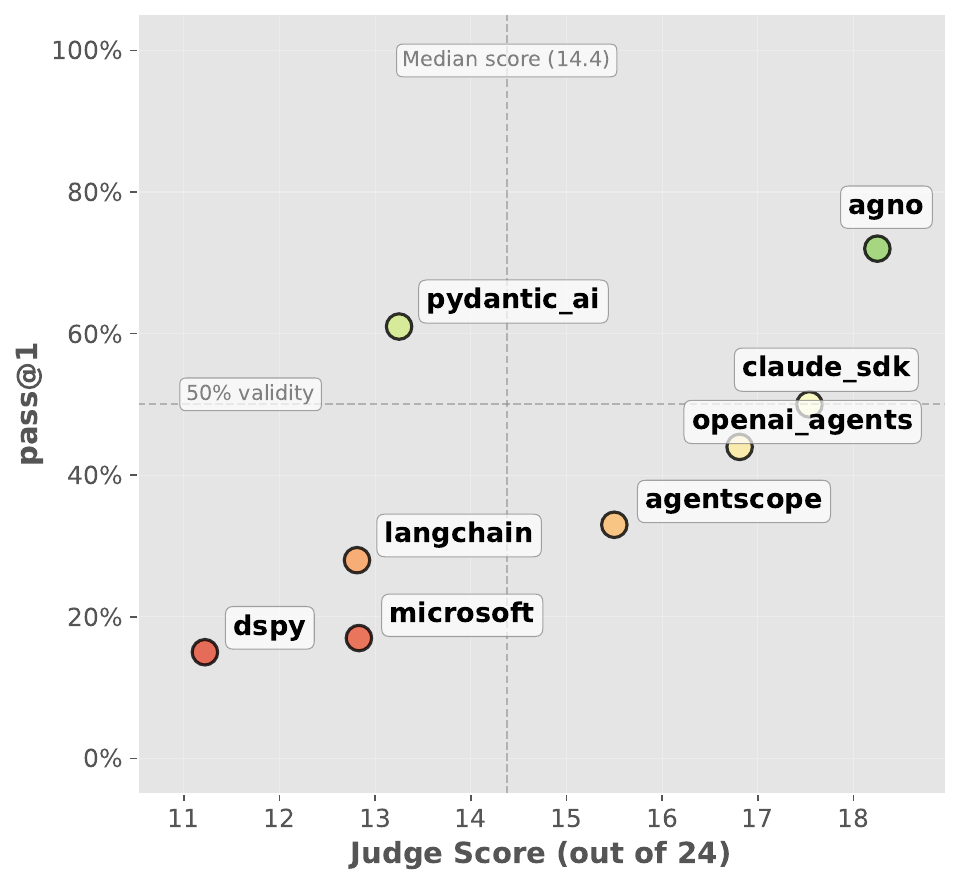}
\end{minipage}
\hfill
\begin{minipage}[c]{0.48\textwidth}
\centering
\small
\begin{tabular}{lrrrr}
\toprule
\textbf{Framework} & \makecell{\textbf{LLOC}\\$(y_f^*)$} & $\bar{\sigma}(f)$ & \textbf{pass@1} & $\mathcal{AI}(f)$ \\
\midrule
Agno            & 54 & 18.29 & 72\% & 0.55 \\
Claude SDK      & 75 & 17.44 & 50\% & 0.36 \\
Pydantic AI     & 52 & 13.18 & 61\% & 0.33 \\
OpenAI Agents   & 70 & 16.92 & 44\% & 0.31 \\
AgentScope      & 82 & 15.72 & 33\% & 0.22 \\
LangChain       & 70 & 12.77 & 28\% & 0.15 \\
Microsoft       & 73 & 12.82 & 17\% & 0.09 \\
DSPy            & 88 & 11.23 & 15\% & 0.07 \\
\bottomrule
\end{tabular}
\end{minipage}
\caption{Framework comparison across judge score and pass@1 dimensions (left) with aggregate AI-assistability rankings (right). The upper-right quadrant represents the \textit{sweet spot}---frameworks where AI assistants reliably generate both structurally aligned and functionally correct implementations.}
\label{fig:framework_scatter}
\end{figure*}

Figure~\ref{fig:framework_scatter} maps each framework along two dimensions: structural alignment (judge score, out of 24) and functional correctness (pass@1). \textbf{Agno} occupies the upper-right \textit{sweet spot} with the highest structural alignment ($\sim$18.3) and pass@1 ($\sim$72\%), demonstrating that AI assistants most reliably generate code for frameworks with a single canonical pattern. \textbf{Pydantic AI} achieves competitive pass@1 ($\sim$61\%) despite a lower judge score ($\sim$13), suggesting that functional correctness can emerge even when generated code diverges from idiomatic patterns. \textbf{Claude SDK} ($\sim$17.5, 50\%) and \textbf{OpenAI Agents} ($\sim$17, 44\%) cluster in the high-alignment, moderate-pass region---structurally sound code that does not always execute successfully.

Below the 50\% validity threshold, \textbf{AgentScope} ($\sim$15.5, 33\%) achieves above-median structural alignment but struggles with runtime correctness. \textbf{LangChain} ($\sim$13, 28\%), \textbf{Microsoft Agent Framework} ($\sim$13, 17\%), and \textbf{DSPy} ($\sim$11, 15\%) form a low-assistability cluster where AI assistants struggle to produce both structurally aligned and functionally correct implementations.

We exclude Google ADK and Smolagents from the $\mathcal{AI}(f)$ ranking---both produce zero valid implementations, so a multiplicative $\mathcal{AI}(f)=0$ regardless of alignment---but treat them as informative failure cases rather than hiding them.\added{ The failures are systematic and reproducible across every run and configuration: Google ADK fails despite reasonable structural alignment due to mandatory session-lifecycle boilerplate, fragmented MCP connection-parameter types, and event-stream interpretation; Smolagents fails on a sync/async execution-model mismatch and unreliable streamable-HTTP MCP support. We give the full technical analysis in Appendix~\ref{appendix:failure_analysis}. Notably, Google ADK strengthens the thesis---convention alignment is necessary but not sufficient when runtime integration complexity overrides structural correctness.}\rc{369D,E}\footnote{Full results including these frameworks are reported in Appendix~\ref{appendix:score_distribution}.}

The composite $\mathcal{AI}(f)$ score (Figure~\ref{fig:framework_scatter}, right) quantifies this landscape. \textbf{Agno} leads decisively ($\mathcal{AI} = 0.55$), nearly $1.5\times$ the next-ranked framework. A clear tier structure emerges: \textbf{Claude SDK} (0.36), \textbf{Pydantic AI} (0.33), and \textbf{OpenAI Agents} (0.31) form a middle tier with comparable overall assistability despite different trade-offs---Claude SDK and OpenAI Agents achieve it through high structural alignment with moderate pass rates, while Pydantic AI compensates for lower alignment with higher functional correctness. Below these, \textbf{AgentScope} (0.22) sits alone as a transitional case, followed by a low-assistability tier where \textbf{LangChain} (0.15), \textbf{Microsoft} (0.09), and \textbf{DSPy} (0.07) score below 0.2, indicating that AI assistants produce usable code in fewer than one in five attempts.

\subsection{Outlier Analysis}
\label{sec:outlier_analysis}

Four frameworks exhibit distinctive patterns revealing the nuanced relationship between framework design and AI-assistability.

\paragraph{Pydantic AI} ($\sim$61\% pass@1, $\sim$13 score) achieves the second-highest functional correctness despite below-median structural alignment. AI-generated code wraps tools with decorators instead of using the direct registration API, diverging from ground-truth patterns yet still producing working implementations.

\paragraph{Claude SDK and OpenAI Agents} ($\sim$50\% and $\sim$44\% pass@1, $\sim$17.5 and $\sim$17 scores) exhibit high structural alignment that does not fully translate to runtime success. Generated code closely follows idiomatic patterns but fails on subtle integration issues such as tool registration ordering and MCP session lifecycle management.

\paragraph{\LangChain} ($\sim$28\% pass@1, $\sim$13 score) suffers from pattern frag\-mentation---its extensive API surface lets AI assistants find \textit{a} pattern without converging on canonical usage. For such multi-pattern frameworks, judge scores underestimate verification burden: functionally correct implementations may rely on deprecated APIs, omit observability hooks, or diverge from team conventions.

\paragraph{DSPy} ($\sim$15\% pass@1, lowest score) lacks native MCP support, forcing improvised integrations via \texttt{httpx} or \texttt{aiohttp}. Despite being the most declarative framework by design, its novel abstractions (signatures, modules, optimizers) are insufficiently represented in AI training data, resulting in the lowest AI-assistability across all frameworks evaluated.


\section{Discussion}
\label{sec:discussion}

\subsection{The Curious Case of \Agno{}}
\label{sec:agno_discussion}

\Agno{} presents a notable paradox: despite having the lowest code complexity (54 LLOC), it is backed by one of the largest library in our study (310K LOC across 3,048 files). Conventional wisdom would suggest such scale increases API surface area and encourages non-idiomatic usage by AI assistants; instead, \Agno{} achieves the highest structural alignment and ties for the best pass@1. This effect is reinforced by its documentation footprint:although substantially larger than \LangChain{}'s (1,429 vs. 762 pages), \Agno{}'s documentation consistently reinforces a single canonical pattern across tutorials and references, enabling AI assistants to converge on the same idiomatic structure. In contrast, \LangChain{} exposes multiple valid but divergent approaches (graph-based, chain-based, and legacy), leading to reasonable execution success (75\% pass@1) but lower structural alignment (15) due to fragmented pattern induction.

This suggests a design principle for AI-assistable frameworks: \textbf{documentation volume amplifies whatever pattern structure exists in the API}. A well-designed API with one obvious way to accomplish tasks benefits from extensive documentation; a flexible API with multiple valid approaches may see documentation exacerbate pattern fragmentation. Framework developers optimizing for AI-assisted development should prioritize API convergence before documentation expansion.

\added{The contrast also clarifies the role of declarative design. \Agno{}'s API is itself declarative, yet convention-aligned, and benefits from that combination; the difficulty \DSPy{} exhibits stems instead from declarative \emph{novelty}---abstractions under-represented in training data, such as its signatures, modules, and optimizers (\S\ref{sec:outlier_analysis}). \Agno{}'s advantage is similarly one of consistency over maturity: although its repository is newer than \LangChain{}'s and \DSPy{}'s, its documentation reinforces a single canonical pattern across tutorials and references.}\rc{369A,C}



\subsection{\added{Controlled Scope and Generalization}}
\label{sec:scope_discussion}

\added{Our findings speak to model developers as much as to framework authors: because AI assistants exploit canonical-pattern alignment, emphasizing a framework's single ``one obvious way''---through fine-tuning or in-context exposure---should improve framework-specific code generation more than exposure to many divergent valid patterns. The evaluation protocol also generalizes beyond this task: substituting a different agentic task, its artifacts, and per-framework ground truth yields the same two-dimensional comparison. The multiplicative form of $\mathcal{AI}(f)$ (\S\ref{sec:copilot}) requires both dimensions to be high and forces $\mathcal{AI}=0$ whenever pass@1$=0$ (e.g.\ Google ADK), and the Agno-first/DSPy-last ordering is stable under additive and weighted-geometric alternatives, so the ranking does not hinge on this combination rule.}\rc{369A,D}

\added{These conclusions hold within a deliberately controlled scope. Fixing the agent architecture, prompts, MCP servers, and entry point leaves the framework integration layer as the only variable, which is what makes $\bar{\sigma}$ measurable; since the Agent-as-a-Tool pattern is expressible in all ten frameworks (verified through the human-authored ground truth), low scores reflect difficulty generating a framework's specific API patterns while the architecture itself remains expressible everywhere. An unconstrained setup that let the assistant choose the architecture, tools, and whether to use a multi-agent system at all would have no shared reference to compare against, collapsing the evaluation to ``does it run?'' The controlled task is thus a prerequisite baseline; unconstrained, agent-chooses-the-structure evaluation is a natural next step we leave to future work.}\rc{369B}

\subsection{LiteLLM: Tooling Effects on AI-Assisted Developer Experience}

While our evaluation focuses on framework-intrinsic properties, we observed that auxiliary tooling can materially shape AI-assisted developer experience. In particular, LiteLLM emerged as a consistent accelerator across frameworks that delegate model management to it. Two features are especially relevant to our findings. First, \textbf{unified model switching} allows developers—and AI coding assistants—to vary providers or model tiers via a single string substitution, enabling rapid development without provider-specific code changes. Second, \textbf{disk-based caching} improves agent-level debugging by replaying cached outputs for unchanged agents, avoiding re-execution of the full pipeline. In our three-tier architecture, this reduced prompt iteration latency from minutes to seconds.

These observations reinforce a broader insight of this work: AI-assistability depends not only on framework APIs and documentation, but on whether the surrounding tooling ecosystem supports fast, deterministic iteration under AI-mediated development. Frameworks that integrate with LiteLLM inherit these benefits transparently, lowering the practical cost of experimentation and error recovery without increasing implementation complexity.



\section{Related Work}
\label{sec:related}

\paragraph{Developer Experience Frameworks.}
\citet{greiler2022actionableframeworkunderstandingimproving} introduce the DX Framework, distilling developer experience (DX) into three core dimensions: \textit{feedback loops} (speed and quality of responses), \textit{cognitive load} (mental effort required), and \textit{flow state} (immersive focus during work). Their study of 21 industry developers identifies 25+ sociotechnical factors affecting DX, with culture, work context, and tooling friction as primary determinants. In conjunction, the SPACE framework~\citep{forsgren2021space} measures productivity across Satisfaction, Performance, Activity, Communication, and Efficiency, showing productivity cannot be reduced to a single metric.

\paragraph{AI Coding Assistant Evaluation.}
\citet{liang2023largescalesurveyusabilityai} survey 410 developers on AI programming assistant usability, finding they value AI assistants for reducing keystrokes and recalling syntax, not brainstorming. They identify \textit{controllability} and \textit{requirement satisfaction} as primary barriers to adoption---developers reject suggestions when generated code misses specifications or resists steering to desired outputs. \citet{ziegler2024copilot} show that acceptance rate of AI-generated suggestions predicts perceived productivity better than persistence or volume metrics, validating self-reported productivity as a meaningful signal. \citet{chen2021evaluating} introduce HumanEval, a benchmark of 164 coding problems to evaluate functional correctness of LLM-generated code, and introduced the pass@k metric now widely used in code generation evaluation. \added{Benchmarks such as HumanEval and SWE-bench~\cite{jimenez2024swebench} measure an \emph{assistant's} general code-generation capability across tasks and repositories; in contrast, $\mathcal{AI}(f)$ measures a \emph{framework's} amenability to AI-assisted generation, holding task and assistant fixed and varying only the framework. The two are complementary: an assistant strong on SWE-bench may still produce brittle code for a framework whose canonical pattern is under-represented in its training data.}\rc{369D,E}

\paragraph{\added{LLM-as-Judge Evaluation.}}
\added{Using a strong LLM as an evaluator has become standard practice for generation quality: LLM judges correlate with human judgments substantially better than reference-overlap metrics (BLEU, ROUGE, BERTScore) for natural-language generation~\cite{liu2023geval} and specifically for code~\cite{zhuo2024icescore}, and checklist-style decomposition of the rubric further improves evaluator agreement~\cite{lee2025checkeval}. We adopt this paradigm for structural alignment (\S\ref{sec:llm_judge}), where the property of interest is structural---tool-registration patterns, MCP lifecycle management, and idiomatic API usage---rather than the token-level surface overlap that embedding-based metrics capture.}\rc{369A,E}

\paragraph{MAF Studies.}
\citet{wang2025empiricalstudyagentdeveloper} present the first large-scale empirical study of agent framework developer practices, analyzing 1,575 GitHub projects and 20,620 developer discussions across ten frameworks. Their findings reveal that 96\% of top-starred projects employ multiple frameworks, suggesting no single framework meets complex real-world needs. Logic failures (33\% of issues), tool integration challenges (25.6\%), and version instability (25\%) dominate reported issues. \citet{duan2024langgraphcrewai} combine LangGraph's graph-based orchestration with CrewAI's role-based collaboration, showing feasibility on email automation and code generation tasks through qualitative case studies.

Our work differs from these prior studies by providing the first controlled, task-based benchmark for systematically comparing MAFs' DX across dimensions of code complexity, and AI-assistability.


\section{Conclusion}
\label{sec:conclusion}

We introduced DDL2PropBank, a controlled benchmark for evaluating multi-agent framework developer experience through the novel task of mapping relational database schemas to PropBank semantic rolesets. By implementing identical agent logic across 10 frameworks, we isolated framework-induced differences in both human implementation effort and AI-assisted code generation. Our evaluation reveals that native MCP integration and unified tool registration are the primary drivers of low code complexity (\S\ref{sec:static_analysis}), while API convergence, not documentation volume, determines AI-assistability (\S\ref{sec:results},\S\ref{sec:discussion}). Structural alignment reliably proxies robustness for single-canonical-pattern frameworks, but underestimates correctness for flexible, multi-pattern MAFs. Among the MAFs evaluated, \Agno{} best exemplifies these design principles, achieving near-lowest complexity and highest AI-assistability ($\mathcal{AI} = 0.55$).\footnote{All artifacts---DDL2PropBank benchmark, PropBank MCP server, and all framework implementations---are available at \url{https://github.com/ahmeshaf/ddl2propbank}. See \S\ref{appendix:artifact_summary} for the full list.}


\section*{Limitations}
\label{sec:limitations}


\paragraph{Task Scope.}
DDL2PropBank is a single instantiation of a broader class of semantic annotation tasks. While the Agent-as-a-Tool pattern we employ is intentionally generic---orchestrator, coordinator, and specialized mappers accessing shared resources---we have not empirically validated that our findings generalize to other annotation tasks such as frame identification or relation extraction. \added{Our claims are accordingly scoped to the Agent-as-a-Tool pattern with shared MCP servers; whether they transfer to graph-based, stateful, long-horizon, or less tool-centric agent systems remains open, and because the task exercises MCP integration heavily, some observed differences reflect MCP-integration maturity alongside core framework design.}\rc{369B,E} Additionally, PropBank is English-centric; multilingual database schemas may require adapted semantic resources.

\paragraph{\added{Evaluation Design.}} \added{Two design choices bound our conclusions. First, $\bar{\sigma}$ is scored against a single human-authored reference per framework; this is intentional---it measures convergence toward canonical API usage (\S\ref{sec:llm_judge})---but it can penalize ``different-but-correct'' implementations in frameworks admitting multiple valid patterns, and a single reference reflects one author's familiarity. Second, the LLM-as-Judge, while grounded in a structured rubric and prior validation of LLM evaluators (\S\ref{sec:llm_judge}), is not validated here against independent human annotation; we report it as a structural proxy rather than a ground-truth quality measure. Multi-reference scoring and human-agreement studies are natural strengthening steps.}\rc{369A,E}

\paragraph{Framework Coverage.}
Our evaluation covers 10 frameworks selected via the criteria in Appendix~\ref{appendix:framework-selection}, but this is not exhaustive. The MAF ecosystem evolves rapidly, with new frameworks emerging and existing ones undergoing significant API changes between versions. Our findings reflect the framework landscape as of late-2025 and may require re-evaluation as the ecosystem matures.

\paragraph{Ephemeral Nature of Evaluation Methodology.}
Our AI-assistability evaluation relies on DDL2PropBank being novel with respect to LLM training data---a property that diminishes as the task becomes widely known. This reflects a broader methodological insight: benchmarks designed to isolate framework quality from model memorization have an inherent shelf life. Reproducing these results in the future may require either continuously novel evaluation tasks or sandboxed experimental setups that prevent training data contamination.

\section*{Future Work}
\label{sec:future}

\paragraph{Ground-Truth Mappings and RelBench Tasks.}
Scaling DDL2Prop\-Bank annotations beyond our current evaluation set requires efficient ground-truth generation. The CoAnnotating methodology \cite{li2023coannotating}, which uses uncertainty-guided work allocation between humans and LLMs, offers a natural fit: by running multiple framework implementations on the same schema, we can treat inter-framework agreement as a confidence signal---high-agreement mappings are automatically accepted while disagreements are routed to human annotators. This approach could efficiently generate ground-truth for the full RelBench benchmark \cite{robinson2024relbench}, which spans 7 databases with 30 prediction tasks, and scale further to REDELEX's 70+ relational databases \cite{peleska2025redelex}. Event-enriched schemas may prove particularly valuable for temporal prediction tasks, where explicit event semantics (e.g., distinguishing \textit{purchase} events from \textit{view} events) could provide interpretable features beyond raw relational structure.

\paragraph{World Simulation and Synthetic Data Generation.}
Generative agent-based modeling enables simulation of human behavior in grounded environments \cite{vezhnevets_generative_2023, wang_simulating_2025}, with applications to post-training data synthesis \cite{tang_synthesizing_2024} and schema-guided event simulation \cite{li_schema-guided_2024}. DDL2PropBank mappings define an action vocabulary for the database schema: PropBank rolesets specify what events each table encodes. Grounding agent actions in these rolesets enables simulation of database operations as semantic events (e.g., users \textit{posting} ads, \textit{requesting} phone numbers). This supports event-driven synthetic data generation for proprietary databases where schemas are available but data access is restricted \cite{balog_user_2025}. Existing LLM-based tabular generation approaches \cite{long_llm-tablogic_2025, xu_are_2024} model statistical distributions without semantic grounding; DDL2PropBank mappings provide event semantics for generating coherent temporal sequences that respect the schema's causal structure.

\paragraph{Benchmark Extension.}
Our evaluation methodology---skeleton code plus documentation-driven generation---is designed for extensibility. As the MAF ecosystem evolves, new frameworks can be added by providing the same skeleton and measuring AI-assistability against the established ground truth. Beyond framework coverage, DDL2PropBank could extend to other semantic resources: FrameNet's situational frames may better capture complex business processes, while VerbNet's syntactic alternations could inform schema normalization. Multilingual database schemas present another frontier, leveraging PropBank's Spanish and Chinese variants or cross-lingual frame resources.


\section*{Ethics Statement}
\label{sec:ethics}

\paragraph{Use of AI in This Work.}
This research employs AI systems at multiple stages: (i) GitHub Copilot generates framework implementations for AI-assistability evaluation (\S\ref{sec:copilot}), (ii) Claude Opus 4.5 serves as an LLM-as-Judge for structural alignment scoring (\S\ref{sec:llm_judge}), and (iii) the DDL2PropBank agents themselves use LLMs for semantic mapping. All AI-generated code and evaluations are disclosed as such; human-authored ground-truth implementations are clearly distinguished from AI-generated variants. The first author used Claude Code to assist in writing the paper (especially in the organization of sections), code development, and experimental analysis, and ChatGPT Deep Research for literature search and fact-checking. While AI tools assisted at various stages, all content has been thoroughly verified by the authors, who take full responsibility for the accuracy of results, validity of claims, and integrity of the work presented.

\paragraph{Risks of AI-Assisted Development.}
Our findings raise a concern for the broader software ecosystem: as AI coding assistants increasingly mediate framework adoption, frameworks optimized for AI-assistability may dominate regardless of their intrinsic merit. This parallels recent findings that ChatGPT decreases idea diversity in brainstorming tasks---while individual outputs may be higher quality, collective diversity suffers when many users converge on similar AI-suggested solutions \cite{meincke2025chatgpt}. In our context, this creates a homogenization risk: if all developers converge on the same ``AI-friendly'' patterns, the diversity of architectural approaches may diminish, potentially foreclosing innovative designs that AI assistants cannot yet generate reliably. Framework developers should balance AI-assistability with support for diverse implementation styles, and the research community should monitor whether AI-assisted development narrows or broadens the space of viable software architectures.

\section*{Acknowledgments}

The puzzle icon in Figure \ref{fig:overview} is sourced from \href{https://www.flaticon.com/free-icon/puzzle-pieces_3050442}{flaticon.com}. We additionally used ChatGPT to generate an AI-rendered variant of the same icon.



\bibliographystyle{ACM-Reference-Format}
\bibliography{refs/datasets,refs/developer_exp,refs/foundational,refs/frameworks,refs/limitations,refs/misc,refs/synth_simulation}

\appendix


\section{Proposition Bank Background}
\label{appendix:propbank}

\textbf{Semantic Role Labeling} (SRL)---the task of identifying predicates in text and labeling their arguments with semantic roles such as agent, patient, and instrument---provides a structured representation of ``who did what to whom'' \cite{gildea2002automatic}. PropBank is a large-scale corpus annotated with predicate-argument structure that serves as the primary training and evaluation resource for SRL systems \cite{palmer2005propbank}.

Relational databases encode the evolving state of a domain through records and the events that modify them. Tables such as \tbl{orders}, \tbl{shipments}, and \tbl{transactions} implicitly represent actions with typed participants: an \tbl{orders} table encodes ordering events where columns like \col{customer\_id} and \col{product\_id} fill semantic roles analogous to PropBank's ARG0 (agent) and ARG1 (theme). This conceptual alignment motivates our use of PropBank as the grounding resource for database schema annotation.

\subsection{From RelBench to Event Semantics}
\label{sec:relbench-motivation}

RelBench \cite{robinson2024relbench} comprises seven real-world relational databases with 30 predictive tasks, all formulated as temporal next-event prediction over user interactions such as clicks, purchases, and content engagement. While these tasks are fundamentally event-driven, relational schemas encode event semantics only indirectly via foreign keys, temporal attributes, and inter-table structure, with event sequences---such as search-to-click or invitation--response--attendance---distributed across multiple tables. Prior relational modeling approaches, such as RelGNN \cite{chen2025relgnncompositemessagepassing}, preserve instance-level structure by operating over populated relational graphs, but require access to underlying data and recover event structure implicitly through representation learning. This leaves a complementary opportunity: deriving explicit event representations directly from schema information alone, providing an interpretable semantic foundation for next-event prediction without relying on instance-level data.

DDL2PropBank bridges this gap by mapping relational database schemas to PropBank semantic rolesets, yielding an explicit and interpretable representation of events directly from schema design. PropBank provides a structured and evolving lexicon of event semantics, in which rolesets define predicate-specific argument structures (e.g., Agent, Theme, Location, Temporal) across a large set of predicates. By aligning database tables with PropBank rolesets, DDL2PropBank transforms multi-table relational structure into explicit event representations that can be queried and reasoned over by downstream systems.

\subsection{Why PropBank?}
\label{sec:why-propbank}

Several lexical-semantic resources could serve as grounding targets, including FrameNet \cite{baker1998framenet}, VerbNet \cite{schuler2005verbnet}, and PropBank \cite{palmer2005propbank}. We select PropBank for four reasons: (i) \textit{scale}---the Unified Verb Index reports 11,637 lemmas with 9,533 rolesets, compared to FrameNet's approximately 800 frames; (ii) \textit{active maintenance}---PropBank is maintained via GitHub (\url{https://github.com/propbank}) with continuous updates through version 3.4 \cite{pradhan2022propbank}; (iii) \textit{cross-resource integration}---SemLink \cite{palmer2009semlink} maps PropBank to VerbNet and FrameNet, while PropBank frames underpin AMR annotation \cite{banarescu2013amr}; and (iv) \textit{domain fit}---PropBank's verb-centric organization and standardized ARG0--ARG4 structure align naturally with event-encoding tables and their participant columns.

\subsection{Roleset Structure}
\label{sec:roleset-structure}

Each PropBank roleset contains the components shown in Table~\ref{tab:roleset-structure}.

\begin{table}[t!]
\caption{PropBank roleset structure.}
\centering
\small
\begin{tabular}{@{}lp{5.5cm}@{}}
\toprule
\textbf{Field} & \textbf{Description} \\
\midrule
\texttt{sense\_id} & Unique identifier (e.g., \texttt{order.02}) \\
\texttt{lemma} & Base verb form (e.g., \texttt{order}) \\
\texttt{definition} & Human-readable meaning (e.g., ``request to be delivered'') \\
\texttt{aliases} & Morphological variants (e.g., \texttt{ordering}) \\
\texttt{roles} & Argument definitions (ARG0, ARG1, etc.) \\
\texttt{examples} & Annotated sentences with argument spans \\
\texttt{lexlinks} & Cross-references to FrameNet, VerbNet \\
\bottomrule
\end{tabular}
\label{tab:roleset-structure}
\end{table}

\paragraph{Core Arguments (ARG0--ARG4).} Numbered arguments are verb-specific but follow proto-role patterns \cite{dowty1991thematic}:

\begin{tabular}{@{}ll@{}}
\textbf{ARG0} & agent, causer, or experiencer \\
\textbf{ARG1} & patient, theme, or entity affected \\
\textbf{ARG2} & instrument, beneficiary, or attribute \\
\textbf{ARG3} & starting point, source, or price \\
\textbf{ARG4} & ending point or destination \\
\end{tabular}

\paragraph{Modifier Arguments (ARGM-*).} Unlike core arguments, modifiers maintain consistent semantics across verbs: ARGM-TMP (temporal), ARGM-LOC (locative), ARGM-MNR (manner), and ARGM-CAU (cause).

\paragraph{Example: \texttt{order.02}.} The roleset ``request to be delivered'' defines four core arguments:
\begin{itemize}[leftmargin=*, nosep]
    \item \textbf{ARG0}: orderer
    \item \textbf{ARG1}: thing ordered
    \item \textbf{ARG2}: benefactive, ordered-for
    \item \textbf{ARG3}: source
\end{itemize}
An annotated example: ``\textit{[Stevie]$_{\text{ARG0}}$ ordered [it]$_{\text{ARG1}}$ [for her]$_{\text{ARG2}}$ [from Mariage Frères]$_{\text{ARG3}}$ [in Paris]$_{\text{ARGM-LOC}}$}.''

\section{PropBank MCP Server}
\label{appendix:propbank_mcp}

\textbf{Model Context Protocol} (MCP)---an open standard for connecting AI assistants to external data sources and tools via JSON-RPC 2.0 message passing---provides a unified interface that decouples agent implementations from tool infrastructure \cite{anthropic2024mcp}. MCP servers expose three primitives: \textit{prompts} (templated instructions), \textit{resources} (readable data), and \textit{tools} (callable functions); clients invoke these primitives through standardized request-response patterns over stdio or HTTP transports.

While LLMs demonstrate impressive language understanding, recent work reveals limitations in structured semantic analysis. \citet{cheng2024srl} show that LLMs struggle with semantic role labeling consistency, particularly for non-core arguments and predicate disambiguation. \citet{ahmed2024xamr} find that downstream performance of symbolic event coreference is bounded by LLM accuracy on semantic role labeling, highlighting SRL as a critical bottleneck. \citet{spaulding2025protoroles} demonstrate that even high-performing models like GPT-4o exhibit degraded accuracy on proto-role property prediction, underperforming human annotators on complex event reasoning.

These findings motivate our PropBank MCP server: rather than relying on LLMs to internalize the full breadth of predicate-argument semantics, we provide programmatic access to PropBank's curated rolesets as an external knowledge source. The MCP architecture enables agents to query structured frame definitions, argument roles, and annotated examples on demand---augmenting LLM reasoning with authoritative lexical-semantic knowledge.

\subsection{Implementation}
\label{sec:mcp-implementation}

We build the PropBank MCP server by cloning the official \texttt{propbank\-frames} repository,\footnote{\url{https://github.com/propbank/propbank-frames}} which contains XML frame files for all PropBank rolesets. We parse these XML files to extract sense definitions, argument structures, aliases, examples, and lexical links to FrameNet and VerbNet. At startup, we construct inverted indexes mapping lemmas and aliases to their corresponding sense identifiers, enabling O(1) lookups for both base verb forms (e.g., ``order'') and morphological variants (e.g., ``ordering'', ``ordered''). The server is implemented using FastMCP,\footnote{\url{https://github.com/jlowin/fastmcp}} a Python framework for building MCP servers, and exposes a StreamableHTTP endpoint that agents can connect to via the standard MCP client protocol.

\subsection{Tool Schemas}
\label{sec:mcp-tools}

The server exposes two tools for querying PropBank:

\paragraph{\texttt{search\_by\_lemma}.} Searches PropBank frames by verb base form or morphological variant.
\begin{itemize}[leftmargin=*, nosep]
    \item \textbf{Parameters}: \texttt{lemma} (str), \texttt{max\_results} (int, default=10)
    \item \textbf{Returns}: List of matching frames with sense\_id, definition, and roles
\end{itemize}

\paragraph{\texttt{search\_by\_sense\_id}.} Retrieves a specific frame by its unique identifier.
\begin{itemize}[leftmargin=*, nosep]
    \item \textbf{Parameters}: \texttt{sense\_id} (str), \texttt{include\_examples} (bool, default=True)
    \item \textbf{Returns}: Complete frame definition including roles, examples, and lexlinks
\end{itemize}

This two-tool design supports the Table Mapper workflow (Algorithm~\ref{alg:tablemapper}): agents first search by lemma to discover candidate rolesets, then retrieve full definitions by sense\_id to examine argument structures and select the best semantic match.
\section{Agent Design}
\label{appendix:agent_design}

\label{appendix:agent_algorithms}



\subsection{Model Configuration}
\label{appendix:model_config}

The starter code pre-specifies model assignments for the DDL2Prop\-Bank agents to ensure consistency across framework implementations:

\begin{itemize}[leftmargin=*, nosep]
    \item \textbf{Orchestrator:} Claude Sonnet 4.5 (high-level workflow coordination)
    \item \textbf{Coordinator:} Claude Haiku 4.5 (filesystem state analysis)
    \item \textbf{Table Mapper:} Claude Haiku 4.5 (per-table PropBank mapping)
    \item \textbf{get\_action\_verbs() tool:} Claude Opus 4.5 (semantic verb generation)
\end{itemize}

This tiered model assignment balances cost and capability: Opus handles the most semantic task (verb generation), Sonnet manages workflow orchestration, and Haiku executes the numerous table mapping operations efficiently. GitHub Copilot must implement the three agent functions using framework-specific APIs while respecting these pre-assigned model configurations.

\subsection{Output Schema}
\label{appendix:output_schema}

Each Table Mapper produces a JSON file conforming to the \texttt{TableMap\-pingOutput} schema:

\begin{lstlisting}[style=yaml, basicstyle=\scriptsize\ttfamily]
TableMappingOutput:
  table_name: string  # e.g., "orders"
  mappings: list[RolesetMapping]

RolesetMapping:
  sense_id: string  # e.g., "order.02"
  lemma: string  # e.g., "order"
  definition: string  # Human-readable meaning
  roles: map[string, string]
    ARG0: string  # Agent/orderer
    ARG1: string  # Patient/thing affected
    ARG2?: string  # Optional: benefactive
    ARG3?: string  # Optional: source
    ARGM-*?: string  # Optional: modifiers
  confidence: float  # 0.0-1.0 semantic fit
\end{lstlisting}

The Coordinator validates mappings against this schema, classifying tables as VALID (non-empty mappings array), MISSING (no JSON file), EMPTY (empty mappings array), or ERROR (malformed JSON).

\subsection{System Prompts}
\label{appendix:system_prompts}

Each ReAct agent prompt follows a three-part structure: (i) \textbf{Preamble}---role description and task context, (ii) \textbf{Available Tools}---MCP tools and Agent-as-a-Tool invocations, and (iii) \textbf{Workflow}---step-by-step execution instructions. The \texttt{GetActionVerbs} LLM-tool differs: as a single-inference wrapper, it has no tool access and uses an \textbf{Instructions} section instead.

\paragraph{Orchestrator.} Top-level coordinator (Figure~\ref{fig:orchestrator_prompt}) with access to Coordinator and Table Mapper as callable tools.

\paragraph{Coordinator and Table Mapper.} Sub-agents (Figure~\ref{fig:agent_prompts}) invoked by the Orchestrator. Coordinator accesses Filesystem MCP; Table Mapper accesses both PropBank and Filesystem MCP.

\paragraph{GetActionVerbs.} LLM-tool (Figure~\ref{fig:getactionverbs_prompt}) that generates candidate action verbs from table context.

\begin{figure}[t!]
\centering
\orchestratorbox{Orchestrator System Prompt}{
\scriptsize\ttfamily
You are an orchestrator agent for PropBank schema mapping. Your job is to coordinate the mapping workflow by invoking specialized tools: the coordinator (for analysis) and the mapper (for creating mappings). \ldots

\textbf{\#\# Available Tools}

\textbf{Coordinator Tool}\\
- \texttt{run\_coordinator()} - Analyzes DDL and checks mapping status \ldots

\textbf{Mapper Tool}\\
- \texttt{run\_mapper(table\_name)} - Maps a single table to PropBank rolesets, outputs \texttt{TableMappingOutput} JSON \ldots

\textbf{\#\# Workflow}

\textbf{Step 1:} Run the coordinator FIRST to understand what work is needed\\
\textbf{Step 2:} Invoke mappers IN PARALLEL for tables that need mapping\\
\textbf{Step 3:} Re-run coordinator to verify completion
}
\caption{System prompt for the Orchestrator agent. The Orchestrator coordinates the mapping workflow by invoking Coordinator and TableMapper as callable tools (Agent-as-a-Tool pattern).}
\label{fig:orchestrator_prompt}
\end{figure}

\begin{figure}[t!]
\resizebox{1\columnwidth}{!}{%
\begin{minipage}[c]{\columnwidth}
  \begin{center}

\promptbox{Coordinator System Prompt}{
\scriptsize\ttfamily
You are a coordinator agent for PropBank schema mapping. \ldots

\textbf{\#\# Available Tools (Filesystem MCP)}

- \texttt{list\_directory(path)} - List files in a directory\\
- \texttt{read\_text\_file(path)} - Read DDL files and JSON mappings

\textbf{\#\# Workflow}

\textbf{Step 1:} Read and parse the DDL file, extract table names\\
\textbf{Step 2:} For each table, check if \texttt{\{table\}.json} exists in output folder\\
\textbf{Step 3:} If exists, validate JSON conforms to \texttt{TableMappingOutput}\\
\textbf{Step 4:} Return list of tables needing mapping (missing or invalid)
}

\vspace{-0.8em}

\promptbox{Table Mapper System Prompt}{
\scriptsize\ttfamily
You are a table mapper agent for PropBank schema mapping. Your job is to map a single database table to PropBank rolesets and output a \texttt{TableMappingOutput} JSON file. \ldots

\textbf{\#\# Available Tools}

\textbf{PropBank MCP}\\
- \texttt{search\_by\_lemma(lemma)} - Search rolesets by verb base form\\
- \texttt{search\_by\_sense\_id(sense\_id)} - Get full roleset definition

\textbf{Filesystem MCP}\\
- \texttt{write\_file(path, content)} - Save \texttt{TableMappingOutput} JSON

\textbf{\#\# Workflow}

\textbf{Step 1:} Analyze table context (columns, PKs, FKs) from DDL\\
\textbf{Step 2:} Generate candidate action verbs via \texttt{GetActionVerbs}\\
\textbf{Step 3:} Search PropBank for matching rolesets\\
\textbf{Step 4:} Map ARG roles to table columns, estimate confidence\\
\textbf{Step 5:} Write \texttt{TableMappingOutput} to output folder
}
\end{center}
\end{minipage}
} 
\caption{System prompts for the Coordinator and Table Mapper sub-agents. Both are invoked by the Orchestrator as callable tools.}
\label{fig:agent_prompts}
\end{figure}

\begin{figure}[t]
\llmtoolbox{GetActionVerbs Prompt}{
\scriptsize\ttfamily
You are a database semantic analysis and PropBank expert. Generate a diverse, non-redundant set of verb candidates that describe the distinct real-world actions recorded by a given database table. \ldots

\textbf{\#\# Instructions}

\textbf{1. Core Action Identification}\\
Ask: What real-world actions does this table RECORD or TRACK?\\
Example: Table \texttt{orders} → order, request, fulfill, cancel, return

\textbf{2. Action Diversity (Avoid Synonyms)}\\
Do NOT include lexical variants (e.g., buy/purchase/acquire).\\
Favor verbs differing along: intent vs execution, initiation vs completion

\textbf{3. Temporal \& Causal Sequencing}\\
Consider: initiation (submit, request) → modification (update) → resolution (complete, cancel)

\textbf{4. Domain-Aware Reasoning}\\
E-commerce → checkout, fulfill, refund | Healthcare → diagnose, prescribe

\textbf{\#\# Output}

Return requested number of verbs as a simple list (base-form, no explanations)
}
\caption{Prompt for the \texttt{GetActionVerbs} LLM-tool. The tool receives table context (name, columns, foreign keys) and returns candidate action verbs for PropBank queries.}
\label{fig:getactionverbs_prompt}
\end{figure}

\section{Framework Feature Comparison}
\label{appendix:framework_comparison}

This appendix provides a detailed feature comparison of the 10 multi-agent frameworks evaluated in this benchmark, along with the rationale for their selection.

\subsection{Selection Criteria}
\label{appendix:framework-selection}

We include all three major LLM provider frameworks---Claude SDK (Anthropic), OpenAI Agents SDK, and Google ADK---regardless of selection criteria, as they represent first-party agentic solutions from the primary model providers. For the remaining seven community-maintained frameworks, we applied eight criteria balancing diversity, maturity, and practical usability:

\begin{itemize}[leftmargin=*, nosep, label={}]
    \item \textbf{C1} (Open-Source): Permissive licenses enabling reproducibility and community scrutiny.
    \item \textbf{C2} (Dual Tool Integration): Support for both MCP servers and direct function calling.
    \item \textbf{C3} (Parallel Tools): Concurrent tool invocations for efficient batch processing.
    \item \textbf{C4} (Observability): Programmatic access to tool counts and token metrics.
    \item \textbf{C5} (Multi-Provider): Support for Anthropic and OpenAI model families.
    \item \textbf{C6} (Single API Key): Operation with one provider's credentials.
    \item \textbf{C7} (Programmatic): CLI or programmatic invocation for batch processing.
    \item \textbf{C8} (Documentation): Comprehensive docs; \texttt{llms.txt} availability valued.
\end{itemize}

\noindent We also prioritized frameworks with significant GitHub stars as a proxy for community validation (Table~\ref{tab:framework-consolidated}).

\begin{table}[t]
\caption{Framework compliance with selection criteria. C1: Open-source, C2: MCP + function calling, C3: Parallel tools, C4: Usage observability, C5: Multi-provider, C6: Single API key, C7: CLI-based, C8: Documentation (\texttt{llms.txt}). \checkmark\ = full support, $\sim$ = partial/manual, -- = not available.}
\label{tab:framework-criteria}
\centering
\small
\begin{tabular}{@{}p{1.5cm}@{\hspace{4pt}}*{8}{c}@{}}
\toprule
\textbf{Framework} & \textbf{C1} & \textbf{C2} & \textbf{C3} & \textbf{C4} & \textbf{C5} & \textbf{C6} & \textbf{C7} & \textbf{C8} \\
\midrule
\multicolumn{9}{@{}l}{\textit{LLM Providers' Frameworks}} \\
\midrule
Claude SDK       & $\sim$ & \checkmark & \checkmark & \checkmark & -- & \checkmark & \checkmark & \checkmark \\
OpenAI Agents    & \checkmark & \checkmark & \checkmark & \checkmark & \checkmark & \checkmark & \checkmark & \checkmark \\
Google ADK       & \checkmark & \checkmark & \checkmark & \checkmark & \checkmark & \checkmark & \checkmark & \checkmark \\
\midrule
\multicolumn{9}{@{}l}{\textit{Independent Frameworks}} \\
\midrule
Pydantic AI      & \checkmark & \checkmark & \checkmark & \checkmark & \checkmark & \checkmark & \checkmark & \checkmark \\
Agno             & \checkmark & \checkmark & \checkmark & \checkmark & \checkmark & \checkmark & \checkmark & \checkmark \\
\LangChain{}     & \checkmark & \checkmark & \checkmark & \checkmark & \checkmark & \checkmark & \checkmark & \checkmark \\
DSPy             & \checkmark & \checkmark & \checkmark & \checkmark & \checkmark & \checkmark & \checkmark & \checkmark \\
Microsoft        & \checkmark & \checkmark & \checkmark & \checkmark & \checkmark & \checkmark & \checkmark & -- \\
Smolagents       & \checkmark & \checkmark & \checkmark & $\sim$ & \checkmark & \checkmark & \checkmark & -- \\
AgentScope      & \checkmark & \checkmark & \checkmark & $\sim$ & \checkmark & \checkmark & \checkmark & -- \\
\bottomrule
\end{tabular}

\end{table}

\subsection{Selected Frameworks}

Table~\ref{tab:framework-consolidated} provides a comprehensive overview of the 10 frameworks included in this benchmark, including their GitHub repositories, documentation URLs, design patterns, and AI-assisted development support (\texttt{llms.txt} availability).

\begin{table*}[t]
\caption{Consolidated framework overview: versions, repositories, documentation, design patterns, and AI-assisted development support. $^\dagger$DSPy uses git commit \texttt{e0833e37} (parallel tool calls branch). $^\ddagger$\LangChain{} uses \texttt{langchain==1.1.3} and \texttt{langgraph==1.0.4}. $^\S$Microsoft Agent Framework version \texttt{1.0.0b251120}.}
\centering
\footnotesize
\resizebox{\textwidth}{!}{
\begin{tabular}{@{}p{2cm}p{1.6cm}rp{2.8cm}p{2.8cm}p{3.2cm}c@{}}
\toprule
\textbf{Framework} & \textbf{Version} & \textbf{Stars} & \textbf{GitHub} & \textbf{Documentation} & \textbf{Design Pattern} & \textbf{llms.txt} \\
\midrule
\multicolumn{7}{@{}l}{\textit{LLM Providers' Frameworks}} \\
\midrule
Claude SDK & 0.1.13 & 2.5K & \href{https://github.com/anthropics/claude-agent-sdk}{anthropics/claude-agent-sdk} & \url{claude.com/docs/agent-sdk} & Native MCP; \texttt{@tool}; built-in usage & \checkmark \\
OpenAI Agents & 0.6.2 & 17.8K & \href{https://github.com/openai/openai-agents-python}{openai/openai-agents-python} & \url{openai.github.io/openai-agents} & Native MCP; \texttt{@function\_tool}; LiteLLM & \checkmark \\
Google ADK & 1.18.0 & 15.6K & \href{https://github.com/google/adk-python}{google/adk-python} & \url{google.github.io/adk-docs} & Native \texttt{McpToolset}; direct func; LiteLLM & \checkmark \\
\midrule
\multicolumn{7}{@{}l}{\textit{Independent Frameworks}} \\
\midrule
Pydantic AI & 1.22.0 & 13.7K & \href{https://github.com/pydantic/pydantic-ai}{pydantic/pydantic-ai} & \url{ai.pydantic.dev} & Native MCP; heterogeneous toolsets; \texttt{.usage()} & \checkmark \\
Agno & 2.3.8 & 35.9K & \href{https://github.com/agno-agi/agno}{agno-agi/agno} & \url{docs.agno.com} & Unified \texttt{MCPTools} (stdio+http); LiteLLM & \checkmark \\
DSPy & e0833e3$^\dagger$ & 30.7K & \href{https://github.com/stanfordnlp/dspy}{stanfordnlp/dspy} & \url{dspy.ai} & External \texttt{mcp} lib; manual tool conversion & \checkmark \\
\LangChain{} & 1.1.3 / 1.0.4$^\ddagger$ & 22.1K & \href{https://github.com/langchain-ai/langgraph}{langchain-ai/langgraph} & \url{langchain-ai.github.io/langgraph} & MCP adapters; \texttt{@tool}; LiteLLM & \checkmark \\
Microsoft Agent Framework & 1.0.0b$^\S$ & 6.3K & \href{https://github.com/microsoft/agent-framework}{microsoft/agent-framework} & \url{learn.microsoft.com/en-us/agent-framework} & Native MCP; provider-specific client & -- \\
Smolagents & 1.23.0 & 23.3K & \href{https://github.com/huggingface/smolagents}{huggingface/smolagents} & \url{huggingface.co/docs/smolagents} & External \texttt{mcp} lib; \texttt{@tool}; LiteLLM & -- \\
AgentScope  \cite{gao2024agentscopeflexiblerobustmultiagent, gao2025agentscope}  & 1.0.10 & 6.4K & \href{https://github.com/modelscope/agentscope}{modelscope/agentscope} & \url{doc.agentscope.io} & Native MCP; provider-specific classes & -- \\
\bottomrule
\end{tabular}
}

\label{tab:framework-consolidated}
\end{table*}

\section{Detailed Code Complexity Analysis}
\label{appendix:code_complexity_analysis}

\begin{table*}[t]
\caption{Complete code complexity metrics for \texttt{database\_mapper.py} implementations, sorted by SLOC within each category after achieving \texttt{pylint} score of 10/10 for each. SLOC/LLOC = source/logical lines of code; CCN = cyclomatic complexity; Cov Stmts/Arcs = coverage statements and arcs (executable paths); Sync/Async Fns = function counts by type; Imports = framework-specific imports. Bold indicates best (lowest) in column. The \textit{Starter (noop)} row shows the skeleton code provided to all implementations as a baseline.}
\centering
\small
\begin{tabular}{@{}lrrrrrrrrrrr@{}}
\toprule
\textbf{Framework} & \textbf{SLOC} & \textbf{LLOC} & \textbf{CCN} & \makecell{\textbf{Avg}\\\textbf{CCN}} & \makecell{\textbf{Cov}\\\textbf{Stmts}} & \makecell{\textbf{Cov}\\\textbf{Arcs}} & \makecell{\textbf{Sync}\\\textbf{Fns}} & \makecell{\textbf{Async}\\\textbf{Fns}} & \makecell{\textbf{Total}\\\textbf{Fns}} & \textbf{Imports} \\
\midrule
\multicolumn{11}{@{}l}{\textit{LLM Providers' Frameworks}} \\
\midrule
OpenAI Agents & 115 & 70 & 14 & 1.75 & 58 & 60 & 5 & 3 & 8 & 5 \\
Google ADK & 126 & 68 & 12 & 3.00 & 61 & 65 & 0 & 4 & 4 & 8 \\
Claude SDK & 139 & 75 & 10 & 2.00 & 63 & 67 & 0 & 5 & 5 & 7 \\
\midrule
\multicolumn{11}{@{}l}{\textit{Independent Frameworks}} \\
\midrule
Pydantic AI & 95 & \textbf{52} & \textbf{3} & \textbf{1.00} & \textbf{46} & \textbf{47} & 0 & 3 & \textbf{3} & 4 \\
Agno & \textbf{92} & 54 & 4 & 1.33 & 49 & 50 & 0 & 3 & \textbf{3} & \textbf{3} \\
\LangChain{} & 115 & 70 & 11 & 2.75 & 61 & 64 & 1 & 3 & 4 & 8 \\
Microsoft & 109 & 73 & 16 & 4.00 & 64 & 72 & 0 & 4 & 4 & 4 \\
Smolagents & 115 & 66 & 15 & 3.00 & 60 & 61 & 4 & \textbf{1} & 5 & 6 \\
AgentScope & 126 & 82 & 10 & 2.00 & 76 & 77 & 2 & 3 & 5 & 10 \\
DSPy & 126 & 88 & 20 & 3.33 & 81 & 85 & 1 & 5 & 6 & 9 \\
\midrule
\textit{Mean} & 118.6 & 71.2 & 12.3 & 2.58 & 63.2 & 66.1 & 1.4 & 3.4 & 4.8 & 6.6 \\
\textit{Std Dev} & 10.9 & 8.8 & 4.3 & 0.84 & 8.8 & 9.5 & 1.7 & 1.1 & 1.3 & 2.0 \\
\midrule
\textit{Starter (noop)} & 42 & 31 & 3 & 1.00 & 28 & 28 & 0 & 3 & 3 & 0 \\
\bottomrule
\end{tabular}

\label{tab:code_complexity_full}
\end{table*}

\subsection{Implementation Pattern Differences}
\label{sec:appendix-pattern-diff}

The complexity metrics in Table~\ref{tab:code_complexity_full} reflect underlying implementation patterns that vary systematically across frameworks. This section presents concrete code excerpts from the ground-truth implementations, illustrating four key patterns that explain the observed complexity differences.

\subsubsection{Tool Registration Patterns}
\label{sec:appendix-tool-reg}

Frameworks differ in whether tools require explicit decorators or can be registered as plain Python functions with schemas inferred from type hints.

\paragraph{Decorator-Based Registration.} Claude SDK, Smolagents, and \LangChain{} require explicit \texttt{@tool} decorators with schema specifications:

\begin{lstlisting}[caption={Claude SDK: Decorator with explicit schema (7 lines overhead)},label={lst:claude-decorator}]
@tool(
    "get_action_verbs",
    "Generate action verb candidates for a table",
    {"table_name": str, "ddl": str, "num_verbs": int},
)
async def get_action_verbs(args: dict) -> str:
    table_name = args["table_name"]
    ddl = args["ddl"]
    num_verbs = args["num_verbs"]
    # ... implementation
\end{lstlisting}

\paragraph{Plain Function Registration.} Pydantic AI and Agno accept plain Python functions, inferring schemas from type hints and docstrings:

\begin{lstlisting}[caption={Pydantic AI: Plain function with inferred schema (0 lines overhead)},label={lst:pydantic-nodec}]
async def get_action_verbs(table_name: str, ddl: str, num_verbs: int) -> str:
    """Get action verbs for a given table.

    Args:
        table_name: Name of the table to analyze
        ddl: DDL schema string
        num_verbs: Number of verbs to generate
    """
    # ... implementation (schema inferred from signature + docstring)
\end{lstlisting}

The decorator requirement adds 5--7 lines of boilerplate per tool function. With the DDL2PropBank architecture requiring 3+ tool registrations, this overhead accumulates.

\subsubsection{Return Type Requirements}
\label{sec:appendix-return-types}

AgentScope uniquely requires all tool functions to return \texttt{ToolResponse} objects wrapping content in \texttt{TextBlock} structures:

\begin{lstlisting}[caption={AgentScope: Required ToolResponse wrapper},label={lst:agentscope-toolresponse}]
from agentscope.tool import ToolResponse
from agentscope.message import TextBlock

def get_action_verbs_tool(table_name: str, ddl: str, num_verbs: int) -> ToolResponse:
    """Wrapper returning ToolResponse instead of plain string."""
    verbs = get_action_verbs(table_name, ddl, num_verbs)
    return ToolResponse(content=[TextBlock(type="text", text=verbs)])
\end{lstlisting}

This pattern requires: (i) additional imports from framework internals, (ii) explicit \texttt{ToolResponse} wrapping for every tool return, and (iii) nested \texttt{TextBlock} construction with type specifications. No other evaluated framework imposes this overhead.

\subsubsection{Token Usage Extraction}
\label{sec:appendix-tokens}

Observability APIs vary from direct attribute access to complex dictionary traversal.

\paragraph{Clean Pattern.} Agno and Pydantic AI expose token counts via direct attribute access:

\begin{lstlisting}[caption={Agno: Direct metrics attribute (2 lines)},label={lst:agno-tokens}]
response = await agent.arun(task_prompt)
in_tokens, out_tokens = response.metrics.input_tokens, response.metrics.output_tokens
\end{lstlisting}

\begin{lstlisting}[caption={Pydantic AI: Method-based access (2 lines)},label={lst:pydantic-tokens}]
response = await agent.run(task_prompt)
usage = response.usage()
print(f"Input: {usage.input_tokens}, Output: {usage.output_tokens}")
\end{lstlisting}

\paragraph{Complex Pattern.} DSPy requires iterating over language model history with defensive checks:

\begin{lstlisting}[caption={DSPy: Manual history parsing (8 lines)},label={lst:dspy-tokens}]
input_tokens, output_tokens = 0, 0
if hasattr(lm, "history") and lm.history:
    for entry in lm.history:
        usage = entry.get("usage", {})
        if usage:
            input_tokens += usage.get("prompt_tokens", 0) or 0
            output_tokens += usage.get("completion_tokens", 0) or 0
return input_tokens, output_tokens
\end{lstlisting}

The 4$\times$ difference in token extraction code (2 vs.\ 8 lines) reflects API design choices that compound across the three agent types in DDL2PropBank.

\subsubsection{Async Compatibility}
\label{sec:appendix-async}

Smolagents executes tools synchronously within an async event loop, requiring the \texttt{nest\_asyncio} workaround:

\begin{lstlisting}[caption={Smolagents: Required event loop patching},label={lst:smolagents-async}]
import nest_asyncio

# Required: enable nested event loops for smolagents compatibility
nest_asyncio.apply()

@tool
def get_action_verbs_tool(table_name: str, ddl: str, num_verbs: int) -> str:
    """Must use sync completion despite async context."""
    response = completion(  # sync version, not acompletion
        model=VERBS_MODEL,
        messages=[{"role": "user", "content": prompt}],
    )
    return response.choices[0].message.content.strip()
\end{lstlisting}

This pattern forces synchronous tool implementations even when async would be natural, introduces potential runtime issues with event loop patching, and limits composability with async-native codebases.

\subsubsection{Pattern Summary}

Table~\ref{tab:pattern-summary} summarizes the implementation pattern differences across frameworks.

\begin{table}[h]
\centering
\resizebox{\columnwidth}{!}{
\begin{tabular}{@{}p{2cm}p{2cm}p{2cm}p{3.5cm}@{}}
\toprule
\textbf{Pattern} & \textbf{Best Case} & \textbf{Worst Case} & \textbf{Differentiator} \\
\midrule
Tool Registration & Pydantic AI, Agno & Claude SDK, Smolagents & Decorator vs.\ plain function \\
Return Wrapping & Most frameworks & AgentScope & Direct return vs.\ ToolResponse \\
Token Extraction & Agno, Pydantic AI & DSPy, Microsoft & Attribute access vs.\ dict parsing \\
Async Compatibility & Most frameworks & Smolagents & Native async vs.\ loop patching \\
\bottomrule
\end{tabular}
}
\caption{Summary of implementation pattern differences across frameworks.}
\label{tab:pattern-summary}
\end{table}
\section{Experimental Setup Details}
\label{appendix:setup_details}

\subsection{CLAUDE.md Project Instructions}
\label{sec:claudemd_appendix}

\texttt{CLAUDE.md} is a project-level instruction file for Claude Code encoding codebase conventions and task-specific requirements. Unlike framework documentation (via \texttt{llms.txt}), it specifies project-specific constraints constant across all implementations. Each AI assistant receives the standardized file (Figure~\ref{fig:claudemd}) containing: (i) \texttt{uv} package manager commands, (ii) Agent-as-a-Tool architecture, (iii) agent hierarchy with tool assignments, and (iv) MCP endpoints. The critical instruction requires LLM orchestration over Python loops---ensuring our evaluation measures framework orchestration capabilities.

\begin{figure}[t!]
\promptbox{CLAUDE.md (Condensed)}{
\scriptsize\ttfamily
\# PropBank SQL Agent\\[0.3em]
Maps database schemas to PropBank rolesets via multi-agent systems.\\[0.4em]
\#\# Dev Commands\\[0.2em]
uv add <package> \# add dependency\\
uv run python main.py --framework <name> \# run agent\\
uv run pytest tests/ \# run tests\\[0.4em]
\#\# Architecture: Agent-as-a-Tool\\[0.2em]
Query -> [Orchestrator] -> [Coordinator/Mapper] x N -> Results\\[0.4em]
\#\# Agent Hierarchy\\[0.2em]
- \textbf{Orchestrator}: workflow (tools: run\_coordinator, run\_mapper)\\
- \textbf{Coordinator}: analyzes DDL (tools: filesystem read-only)\\
- \textbf{Mapper}: maps tables (tools: filesystem, PropBank MCP)\\[0.4em]
\#\# Critical Requirement\\[0.2em]
\textbf{Use LLM orchestration, NOT Python loops.}\\
Orchestrator must decide workflow via tool calls, not asyncio.gather().\\[0.4em]
\#\# MCP Servers\\[0.2em]
PropBank: search\_by\_lemma(), search\_by\_sense\_id()\\[0.4em]
\#\# Task\\[0.2em]
Complete \texttt{database\_mapper.py}. Do NOT edit main.py or prompts.py.
}
\caption{Condensed \texttt{CLAUDE.md} provided to AI assistants, encoding dev commands, architectural constraints, and the LLM orchestration requirement.}
\label{fig:claudemd}
\end{figure}

\subsection{Test Database}
\label{appendix:test_database}

We use the \textbf{rel-avito} database from RelBench \citep{robinson2024relbench} as the evaluation schema. This classified advertisements dataset contains 8 tables with realistic foreign key relationships:

\begin{itemize}[leftmargin=*, nosep]
    \item \textbf{Core entities:} \tbl{Users}, \tbl{Categories}, \tbl{Locations}
    \item \textbf{Listing lifecycle:} \tbl{Ads}, \tbl{AdInfo}, \tbl{ItemInfo}
    \item \textbf{User interactions:} \tbl{SearchInfo}, \tbl{PhoneRequests}
\end{itemize}

This schema exercises all DDL2PropBank components: the Coordinator must enumerate 8 pending mappings, Table Mapper agents must handle varying table complexities (2--9 columns), and the Orchestrator must coordinate parallel mapping operations. The schema's mix of entity tables (\tbl{Users}, \tbl{Categories}) and event tables (\tbl{PhoneRequests}, \tbl{SearchInfo}) provides diverse semantic grounding challenges.

\subsection{\added{Deployment Complexity}}
\label{appendix:deployment_complexity}

\added{Deploying DDL2PropBank to evaluate a framework is intentionally lightweight, lowering the barrier to extending the benchmark. Setup requires three steps: (i) launch the PropBank MCP server (\texttt{pip install} then \texttt{uvicorn}); (ii) launch the filesystem MCP server (via \texttt{npx}); and (iii) run \texttt{main.sh} with the target framework and database arguments. End-to-end setup completes in under ten minutes on a typical developer machine. Adding a new framework requires only completing the shared \texttt{database\_mapper.py} skeleton (Listing~\ref{lst:skeleton}) against the same fixed infrastructure---no changes to prompts, MCP servers, or the entry point---so the marginal cost of evaluating an additional framework is small.}\rc{369C}

\subsection{Code Skeleton}
\label{sec:skeleton_appendix}

Listing~\ref{lst:skeleton} shows the shared skeleton structure for \texttt{database\_mapper.py} that AI assistants must complete. The skeleton provides function signatures, docstrings, and commented instructions indicating the expected implementation steps for each agent function.

\begin{lstlisting}[language=Python,caption={Shared skeleton for \texttt{database\_mapper.py} across all frameworks},label={lst:skeleton}]
async def run_orchestrator(ddl_file, db_name, output_folder, max_rolesets_per_table):
    """Run the full PropBank mapping workflow."""

    async def run_coordinator() -> str:
        """Run coordinator to get tables needing mapping."""
        # 1. Create coordinator agent with filesystem MCP toolset
        # 2. Run the agent to get the coordinator report
        # 3. Print token usage (input_tokens and output_tokens)
        coordinator_task_prompt = build_coordinator_prompt(...)
        ...

    async def run_mapper(table_name: str, ddl: str) -> str:
        """Map a single table to PropBank rolesets."""
        # 1. Create mapper agent with filesystem MCP, propbank MCP, verbs function
        # 2. Run the agent to save mapping to output_folder/{db_name}/{table_name}.json
        # 3. Print token usage (input_tokens and output_tokens)
        mapper_prompt = build_mapper_prompt(...)
        ...

    # Orchestration logic:
    # 1. Create orchestrator agent with filesystem MCP toolset
    # 2. Register coordinator and mapper functions as tools (Agent-as-a-Tool)
    # 3. Run orchestrator: coordinator finds tables, mapper processes each
    # 4. Print token usage and prepare final report
    orchestrator_task_prompt = build_orchestrator_prompt(...)
    ...
\end{lstlisting}

\section{Fine-grained Analysis of AI-Assistability}
\label{appendix:results_analysis}

\subsection{Score Distribution and Validity Correlation}
\label{appendix:score_distribution}

\begin{figure}[t!]
\centering
\includegraphics[width=0.75\linewidth]{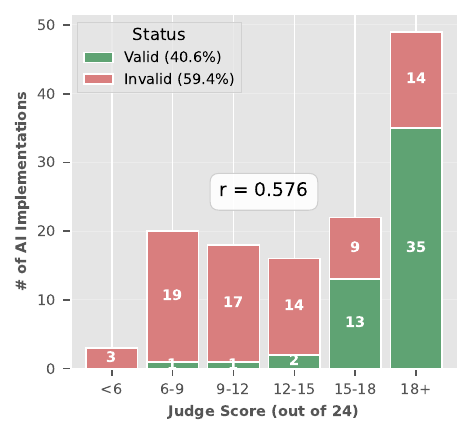}
\caption{Distribution of implementations by judge score and pass@1 status. Higher scores strongly predict runtime success: the 18+ bin achieves 89\% pass@1 (24/27) while the 6--9 bin achieves 0\% (0/4). The Pearson correlation (\added{$r = 0.565$ on the GitHub Copilot subset; $r = 0.576$ across all three assistants, \S\ref{sec:overall_results}}\rc{369E}) validates structural alignment as a meaningful proxy for functional correctness.}
\label{fig:judge_stacked_bars}
\end{figure}

Figure~\ref{fig:judge_stacked_bars} shows the joint distribution of LLM-as-Judge scores and pass@1 outcomes across all framework configurations. The stacked bars partition implementations into passing (green) and failing (red) categories within each score bin. Higher structural alignment scores strongly predict runtime success: implementations scoring 18+ achieve 89\% pass@1 (24/27), the 15--17 bin achieves 73\% (11/15), and scores below 12 yield only 29\% (2/7). The 6--9 bin contains zero passing implementations (0/4), confirming that low structural alignment reliably indicates functional failure.

\subsection{Cost-Effectiveness Analysis}

Table~\ref{tab:cost_effectiveness} reports inference costs for ground-truth implementations processing the rel-avito database. We measure total API cost, rolesets discovered, and cost per roleset—a proxy for semantic work efficiency.

\begin{table}[t!]
\caption{Cost-effectiveness of ground-truth implementations on rel-avito. \textbf{RS}: total rolesets found. \textbf{HC}: high-confidence rolesets. \textbf{Cost}: total API cost (USD). \textbf{\$/RS}: cost per roleset discovered. Frameworks below the rule are 3--5$\times$ more expensive per roleset.}
\centering
\resizebox{0.9\columnwidth}{!}{
\begin{tabular}{@{}lrrrrr@{}}
\toprule
\textbf{Framework} & \textbf{Time (s)} & \textbf{RS} & \textbf{HC} & \textbf{Cost} & \textbf{\$/RS} \\
\midrule
Claude SDK & 508.8 & 119 & 82 & \$1.24 & 0.010 \\
LangGraph & 122.3 & 120 & 75 & \$1.32 & 0.011 \\
Agno & 155.4 & 113 & 86 & \$1.50 & 0.013 \\
Google ADK & 125.6 & 117 & 76 & \$1.57 & 0.013 \\
DSPy & 168.2 & 116 & 78 & \$1.62 & 0.014 \\
Smolagents & 138.1 & 119 & 63 & \$2.07 & 0.017 \\
OpenAI Agents & 167.1 & 124 & 84 & \$2.37 & 0.019 \\
\midrule
Pydantic AI & 203.7 & 124 & 90 & \$6.33 & 0.051 \\
Microsoft & 798.0 & 108 & 80 & \$6.19 & 0.057 \\
\bottomrule
\end{tabular}
}
\label{tab:cost_effectiveness}
\end{table}

\subsection{Library Size and Documentation Coverage}
\label{appendix:library_size}

Table~\ref{tab:library_size} compares library complexity (source lines of code) with documentation coverage (doc page count) for three frameworks with available data. Despite having the largest codebase, \Agno{} provides the most extensive AI-friendly documentation---nearly twice that of \LangChain{}. This extensive yet consistent documentation may contribute to \Agno{}'s superior structural alignment: AI assistants encounter the same canonical patterns across 1,429 pages, reinforcing idiomatic usage rather than fragmenting attention across competing approaches.

\begin{table}[t!]
\caption{Framework library size and documentation coverage. Code LOC measured via \texttt{cloc} on Python files. *\LangChain{} combines \texttt{langchain} (203K) and \texttt{langgraph} (79K) repositories; doc pages combine \texttt{docs.langchain.com} (622) and \texttt{langgraph} docs (140). \added{Pass@1 here is the GitHub Copilot subset; all-assistant aggregates are in Figure~\ref{fig:framework_scatter}.}}
\label{tab:library_size}
\centering
\begin{tabular}{@{}lrrrc@{}}
\toprule
\textbf{Framework} & \shortstack{\textbf{Lib}\\\textbf{Files}} & \shortstack{\textbf{Lib}\\\textbf{LOC}} & \shortstack{\textbf{Doc}\\\textbf{Pages}} & \textbf{Pass@1} \\
\midrule
\Agno{} & 3,048 & 309,799 & 1,429 & 83\% \\
\LangChain{}* & 2,511 & 282,098 & 762 & 75\% \\
\PydanticAI{} & 403 & 130,089 & 127 & 67\% \\
\bottomrule
\end{tabular}

\end{table}

\subsection{\added{Failure Analysis: Google ADK and Smolagents}}
\label{appendix:failure_analysis}

\added{Google ADK and Smolagents are excluded from the $\mathcal{AI}(f)$ ranking (\S\ref{sec:framework_landscape}) because they produce zero valid implementations across all configurations---a systematic, reproducible failure that recurs identically across runs. We retain them in the structural-alignment and complexity analyses and document the failure modes here, as they are informative data points rather than artifacts to hide. Neither failure is explained by structural alignment alone, motivating the tempered reading of the $\bar{\sigma}$--pass@1 correlation (\S\ref{sec:overall_results}).}\rc{369D,E}

\paragraph{\added{Google ADK}} \added{fails despite reasonable structural alignment, driven by runtime-integration complexity that assistants cannot reliably assemble: (i) mandatory session-lifecycle boilerplate (\texttt{InMemorySessi\-onService} plus a \texttt{Runner}) with no counterpart in convention-aligned frameworks; (ii) fragmentation across multiple MCP connect\-ion-parameter types, where the correct one must be selected; and (iii) event-stream interpretation, where results are extracted via conditional logic over a streamed event sequence rather than a direct return value. Each requirement is individually manageable, but their combination consistently yields non-executing implementations---strengthening the thesis that convention alignment is necessary but not sufficient when runtime integration complexity dominates.}\rc{369D,E}

\paragraph{\added{Smolagents}} \added{fails on an execution-model mismatch: its synchronous \texttt{ToolCallingAgent} conflicts with the asynchronous evaluation harness (requiring a \texttt{nest\_asyncio} workaround assistants rarely apply), tool definitions are coupled to \texttt{ToolCollection} context managers, and its streamable-HTTP MCP support did not work reliably in our setup, pushing assistants toward manual \; \texttt{requests.post()} workarounds. These barriers originate in the frameworks themselves under the Agent-as-a-Tool + MCP workload.}\rc{369E}

\section{Artifact Summary}
\label{appendix:artifact_summary}

Table~\ref{tab:artifacts} summarizes the artifacts released with this work. All code will be released under the CC BY-NC-ND license. The PropBank MCP server data is derived from PropBank’s existing annotations and is distributed under the CC BY-SA 4.0 license. The \textbf{rel-avito} database from RelBench is used under the MIT license.

\begin{table*}[t!]
\caption{Summary of released artifacts. The complete benchmark is available at \url{https://github.com/ahmeshaf/ddl2propbank}.}
\centering
\resizebox{0.9\textwidth}{!}{
\small
\begin{tabular}{@{}p{2.5cm}p{3.5cm}p{9cm}@{}}
\toprule
\textbf{Artifact} & \textbf{Location} & \textbf{Description} \\
\midrule
\multicolumn{3}{@{}l}{\textit{Core Benchmark}} \\
\midrule
Task Definition & \textit{This paper} & DDL2PropBank specification and evaluation rubric \\
Test Databases & \texttt{data/*.sql} & 7 RelBench schemas in DDL format \\
Example Outputs & \texttt{output/rel-avito/*/} & Schema-to-roleset mappings by each framework for rel-avito database schema DDL for a maximum of 15 roleset mappings per table. \\
\midrule
\multicolumn{3}{@{}l}{\textit{MCP Server}} \\
\midrule
PropBank MCP & \texttt{mcp/propbank-mcp/} & \texttt{search\_by\_lemma}, \texttt{search\_by\_sense\_id} tools for programmatic PropBank access \\
\midrule
\multicolumn{3}{@{}l}{\textit{Framework Implementations}} \\
\midrule
Ground-Truth Code & \texttt{gt/*/} & 10 human-authored framework implementations \\
AI-Generated Code & \texttt{src/*/} & 60 Copilot implementations (10 frameworks $\times$ 6 configs) \\
Starter Template & \texttt{starter/} & Shared infrastructure and MCP setup \\
System Prompts & \texttt{starter/prompts.py} & Orchestrator, Coordinator, and Table Mapper prompts \\
\midrule
\multicolumn{3}{@{}l}{\textit{Evaluation Tools}} \\
\midrule
LLM-as-Judge & \texttt{eval/llm\_judge.py} & Structural alignment prompts and scoring scripts \\
Functional Analysis & \texttt{eval/func\_analysis.py} & Three-dimensional analysis of pass@1, judge scores, and complexity \\
Static Analyzer & \texttt{eval/cc.py} & LLOC, CCN extraction scripts \\
\bottomrule
\end{tabular}
}
\label{tab:artifacts}
\end{table*}

\end{document}